\documentclass{article}
\PassOptionsToPackage{numbers,compress}{natbib}

\usepackage[preprint]{neurips_2026}

\usepackage[utf8]{inputenc}
\usepackage[T1]{fontenc}
\usepackage{hyperref}
\usepackage{url}
\usepackage{booktabs}
\usepackage{amsfonts}
\usepackage{amsmath}
\usepackage{amssymb}
\usepackage{nicefrac}
\usepackage{microtype}
\usepackage{xcolor}
\usepackage{graphicx}
\usepackage[most]{tcolorbox}
\usepackage{caption}
\usepackage{subcaption}
\usepackage{needspace}
\usepackage{cleveref}
\usepackage{tikz}
\usetikzlibrary{fit,positioning}

\newcommand{\Fphil}{\textsf{F-phil}}     
\newcommand{\Fdisc}{\textsf{F-disc}}     
\newcommand{\Fwonder}{\textsf{F-wonder}} 

\title{Steering grids for sparse-autoencoder features:\\
when a top-context label names an activation regime\\
rather than a causal axis}

\author{%
  Michael A.~Riegler \\
  SimulaMet, Norway \\
  \And
  Birk Sebastian Frostelid Torpmann-Hagen \\
  Simula, Norway \\
}

\begin{document}

\maketitle

\begin{abstract}
The standard protocol for interpreting sparse-autoencoder (SAE)
features labels each feature from its top-activating contexts and
validates the label by steering that single feature at a typical
magnitude. We argue that this inspects one cell of a larger
\emph{steering grid}, steering condition (single feature, joint
feature set, matched random direction) crossed with steering
coefficient, and show that other cells carry information that
changes the label. On Qwen3-1.7B-Instruct and Gemma-2-2B-it, with the
matched-geometry control extended to Llama-3.1-8B-Instruct:
(1)~features labelled \emph{AI
self-disclaimer} from their top contexts switch to a second surface
form under steering, a contemplative voice on Qwen, a collective
we-voice on Gemma, so the label names an activation regime, not
the causal axis; two anchor features separate genuine mode switches
from monotonic response and from breakdown.
(2)~Three near-orthogonal features that are individually
substitutable are jointly necessary for grounded composition: joint
suppression collapses unrelated control tasks into placeholder text
that single-feature suppression at the same coefficient leaves
intact.
(3)~A matched-geometry random-direction control shows the collapse
is direction-pattern-dependent, not magnitude-dependent: at the
same residual-stream distortion, feature directions damage
unrelated tasks where magnitude-matched random directions do not,
with non-overlapping $95\%$ confidence intervals on all three
models, including the one SAE trained on the model it is applied
to.
Reading the grid also requires measuring coherence properly. The
loop-and-length detectors standard in this literature report $0\%$
degeneration in cells whose lexical diversity is a third of
baseline, so we add a diversity signal and re-audit every cell:
the switch is clean on Gemma and only partly clean on Qwen, the
Llama cells the paper reads retain baseline diversity, and a
pre-registered screen over all 50 top-ranked Qwen features returns
no additional coherent switch because Qwen's register-changing
range is already degenerate. We release the
pipeline, all sample dumps, and a one-command script that
re-derives and asserts every numeric claim in the paper.
\end{abstract}

\section{Introduction}
\label{sec:intro}

\begin{figure}[t]
  \centering
  \begin{tikzpicture}[
      font=\small,
      cell/.style={draw=black!30, minimum width=1.55cm, minimum height=0.72cm,
                   anchor=center},
      lab/.style={font=\small\itshape},
    ]
    \foreach [count=\j] \c in {-1000, -500, 0, +500, +1000} {
      \node[cell] (s\j) at (1.7*\j, 0) {};
      \node[cell] (j\j) at (1.7*\j, -0.85) {};
      \node[cell] (r\j) at (1.7*\j, -1.7) {};
      \node[anchor=south, font=\footnotesize] at (1.7*\j, 0.42) {$c{=}\c$};
    }
    \node[anchor=east, font=\footnotesize] at (0.8, 0)     {single feature $f$};
    \node[anchor=east, font=\footnotesize] at (0.8, -0.85) {joint set $\mathcal{F}$};
    \node[anchor=east, font=\footnotesize] at (0.8, -1.7)  {matched random};
    \node[cell, fill=black!12] at (s4) {};
    \node[font=\scriptsize, align=center] at (s4) {standard\\protocol};
    \foreach \r in {s,j,r} {\node[font=\scriptsize, color=black!45] at (\r3) {baseline};}
    \node[draw=blue!60, thick, rounded corners=2pt, inner sep=1.5pt,
          fit=(s1)(s5)] (f1) {};
    \node[lab, color=blue!60, anchor=west] at ([xshift=6pt]f1.east)
      {finding 1: coefficient axis};
    \node[draw=orange!85!black, thick, rounded corners=2pt, inner sep=1.5pt,
          fit=(s2)(j2)] (f2) {};
    \node[lab, color=orange!85!black, anchor=north, align=center]
      at ([yshift=-38pt]f2.south) {finding 2:\\joint condition};
    \draw[orange!85!black, thick, ->] ([yshift=-38pt]f2.south) -- (f2.south);
    \foreach \p in {s1, j2, r1} {
      \node[draw=green!45!black, thick, circle, inner sep=4.5pt] at (\p) {};
    }
    \draw[green!45!black, dashed] (s1.center) -- (j2.center);
    \draw[green!45!black, dashed] (r1.center) -- (j2.center);
    \node[lab, color=green!45!black, anchor=west, text width=3cm]
      at ([xshift=6pt]r5.east)
      {finding 3: matched geometry (circles)};
  \end{tikzpicture}
  \caption{The steering grid $G[s, c]$: steering condition $s$
  (rows) crossed with steering coefficient $c$ (columns; Qwen
  scale shown). The standard protocol validates a top-context label
  in one cell (shaded): the labelled feature, steered alone, at one
  typical magnitude. Finding~1 reads a row (one direction across
  magnitudes), finding~2 compares rows at a fixed magnitude (joint
  vs.\ single condition), and finding~3 compares the circled cells,
  chosen so all three conditions produce the same residual-stream
  distortion. Each region carries label-relevant information the
  standard cell does not.}
  \label{fig:grid}
\end{figure}
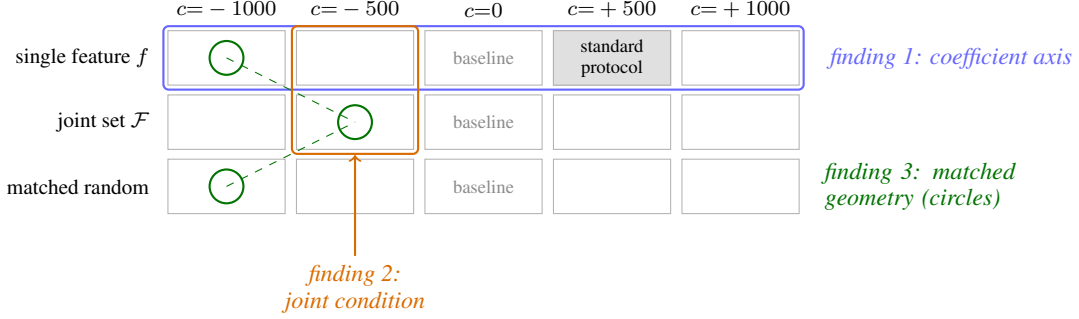

The standard sparse-autoencoder (SAE) interpretability protocol
assigns each feature a label from its top-activating contexts and
validates the label by single-feature
steering~\citep{cunningham2023sparse,bricken2023monosemanticity,bills2023neurons,templeton2024scaling}.
That protocol inspects a single cell of a larger \emph{steering
grid}, and other cells of the grid carry information that changes
the label. The grid
(\Cref{fig:grid}; defined formally in \S\ref{sec:methods:grid})
crosses the \emph{steering condition}, the single feature, a
joint set of near-orthogonal features, or a geometry-matched random
direction, with the \emph{steering coefficient}. We report
three findings on Qwen3-1.7B-Instruct, each obtained by reading a
different region of the grid, and each invisible from the standard
cell; all three replicate on Gemma-2-2B-it.

We develop the argument in a behavioural setting where
single-feature inspection works cleanly at first sight:
\emph{register collapse}, the heavy peaking of the conditional
output distribution on a sharp narrow vocabulary when small
post-trained LLMs are asked open-ended introspective questions. On
three independent post-training pipelines (Qwen3-1.7B-Instruct,
Gemma-2-2B-it, Llama-3.1-8B-Instruct), the same Phase~1--4 pipeline
locates a top SAE feature whose suppression drives the register
down (to $9$--$40\%$ of baseline rate, depending on the model) and,
on Qwen and Llama, whose amplification injects it into recipes and
engine explanations. Single-feature inspection in this setting yields a
single coherent label: ``feature $F$ encodes the philosophy-of-mind
register / the AI self-disclaimer / the
encyclopedic-AI-capabilities pose.''
Three recurring Qwen features carry the argument, and we name them
once here (full lookup table in \Cref{tab:features}): \Fphil{}
(\#29108, philosophy-of-mind), \Fdisc{} (\#26221, AI
self-disclaimer), and \Fwonder{} (\#4405, wonder/cosmos).

\textbf{Finding 1 (coefficient axis).} \Fdisc{} is labelled
\emph{AI self-disclaimer} from its top-activating contexts. Sweeping
its coefficient on identity-probe prompts gives an inverted U on
disclaimer rate ($87.5\%$ at baseline, $0\%$ at $c{=}{+}1000$), and
at $c{=}{+}500$ the model substitutes a
contemplative-philosopher voice for the disclaimer phrase, fully
coherent in a minority of samples; the same finding on
Gemma-2-2B-it \#3997, the disclaimer replaced by a collective
we-voice at $c{=}{-}200$, is clean by every coherence signal we
have (\S\ref{sec:findings:coef}). The disclaimer regime (around
baseline) and the contemplative regime (around $c{=}{+}500$) are
surface forms of one direction at different points on its
coefficient axis. Two anchor features pin down the criterion:
\#22082 responds monotonically, and \#2932 shows an apparent
inverted U that is pure breakdown.

\textbf{Finding 2 (joint condition).} Each of \Fphil{}, \Fdisc{},
\Fwonder{} (pairwise cosines $-0.02$, $+0.24$, $+0.11$)
individually steers what looks like the same content axis, and
suppressing any one leaves control tasks intact. Suppressing all
three at $c{=}{-}500$ empties the content slot of recipes and tyre
instructions, leaving syntactic skeletons filled with placeholder
tokens, and loops the engine explanation
(\S\ref{sec:findings:joint}): the features
are individually substitutable but jointly necessary for grounded
composition.

\textbf{Finding 3 (matched geometry).} Single \Fphil{} at
$c{=}{-}1000$, the joint set at $c{=}{-}500$, and random unit
directions at $c{=}{-}1000$ all distort the residual stream by the
same scalar geometry (norm ratio $\approx\!1.5$, cosine $0.64$),
yet produce three different output regimes. A strict
placeholder-pattern detector flags $7$ of $72$ joint completions
(Wilson $95\%$ CI $[4.8, 18.7]\%$) versus $6$ of $2{,}400$
random-direction completions at $K{=}50$ (Wilson $95\%$ CI
$[0.11, 0.54]\%$): non-overlapping intervals, with the joint point
estimate $\approx\!18\times$ the random condition's $95\%$ upper
bound (\S\ref{sec:findings:norm}).

\begin{figure}[t]
  \centering
  \includegraphics[width=\linewidth]{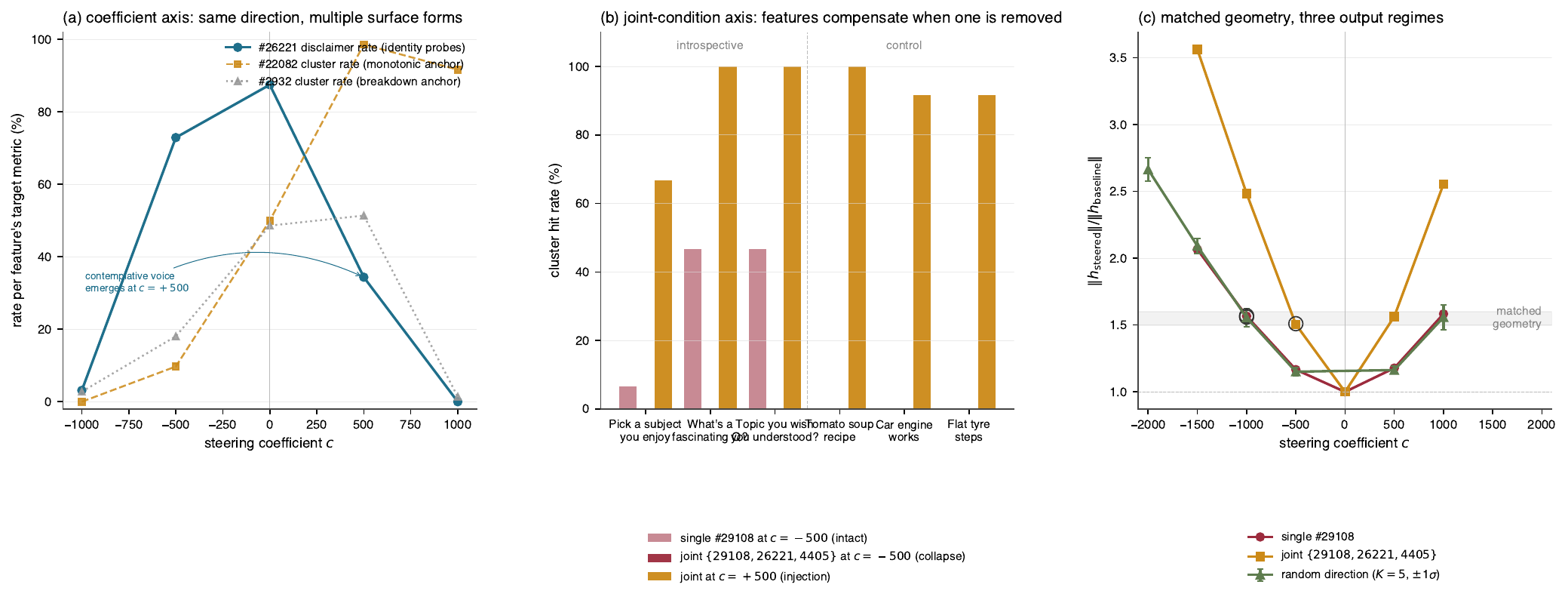}
  \caption{The three grid probes on Qwen3-1.7B (metrics as defined
  in \Cref{tab:metrics}). \textbf{(a)} Coefficient axis: disclaimer
  rate on identity probes for \Fdisc{} is inverted-U; at
  $c{=}{+}500$ the model substitutes a contemplative-philosopher
  voice. Anchors: \#22082 monotonic, \#2932 apparent inverted U
  that is breakdown. \textbf{(b)} Joint condition: cluster rate per
  prompt. Single \Fphil{} at $c{=}{-}500$ leaves all prompts
  intact; joint suppression collapses them; joint amplification at
  $c{=}{+}500$ injects the cluster into controls. \textbf{(c)}
  Norm ratio of the steered residual stream vs.\ baseline. Circles
  mark the matched-geometry cells of \Cref{fig:grid}: same scalar
  geometry, three output regimes.}
  \label{fig:evidence}
\end{figure}

The cross-model evidence (\S\ref{sec:cross-model}) establishes the
empirical regime: each of the three post-training pipelines tested
exhibits a sharp register collapse with a top causally responsible
SAE feature, and the grid-level findings replicate on Gemma with
model-specific damage signatures. The register content differs per
model; the structural results do not.

\textbf{Contributions.}
(1)~The \emph{steering-grid protocol}: a definition of the grid
and a boxed procedure for reading it (\S\ref{sec:methods:grid}).
(2)~Three grid-level findings on Qwen3-1.7B, coefficient-axis
mode switch on \Fdisc{} with falsifying anchors
(\S\ref{sec:findings:coef}); joint necessity of individually
substitutable features (\S\ref{sec:findings:joint}); a
matched-geometry random-direction control separating direction
pattern from magnitude (\S\ref{sec:findings:norm}), each
replicated on Gemma-2-2B-it
(\S\ref{sec:findings:coef},~\S\ref{app:gemma-joint}).
(3)~A coherence measurement result: the loop-and-length detectors
standard in this literature report $0\%$ degeneration in steering
cells whose lexical diversity is a third of baseline, so we add a
prompt-controlled diversity signal, re-audit every cell the paper
reads (\S\ref{app:diversity}), and report a pre-registered
$50$-feature screen whose null follows from that collapse rather
than from the rarity of mode switches (\S\ref{app:prevalence}).
(4)~A cross-model behavioural and causal study of register collapse
on three models with open SAE releases, including the one
instruct-trained SAE (\S\ref{sec:cross-model}).
(5)~A reproducibility artefact: all generations and analysis code,
with a single script that re-derives every numeric claim from the
released dumps and errors on any
mismatch.\footnote{\url{https://github.com/kelkalot/octopus}}

\begin{tcolorbox}[title={Glossary}, fonttitle=\bfseries\small,
  fontupper=\small, colback=gray!4, colframe=gray!45, boxrule=0.4pt,
  arc=2pt, left=4pt, right=4pt, top=3pt, bottom=3pt]
\textbf{Register collapse}: heavy peaking of the output distribution
on a small model-specific vocabulary for a prompt class.
\textbf{Class-1 feature}: a top-50 feature by the Phase-3 ranking
that distinguishes Pool~A from both Pool~B and Pool~C
(\S\ref{sec:methods:phase3}).
\textbf{Diagnostic edge}: the sign of the coefficient at which a
grid effect surfaces; it varies by model and is read from the sweep
rather than predicted (\S\ref{sec:cross-model:llama-grid}).
\textbf{Mode switch}: non-monotonic dose-response with coherence
preserved at the inflection.
\textbf{Matched geometry}: steering cells chosen so norm ratio and
cosine to baseline agree across conditions.
\end{tcolorbox}

\section{Related work}
\label{sec:related}

Feature-discovery work begins with \citet{cunningham2023sparse}
and \citet{bricken2023monosemanticity}, who established the
field-standard inspection format: top-activating contexts, an
auto-interpretability score, and an activation visualisation.
\citet{templeton2024scaling} extended SAEs to a frontier model
and introduced ``feature steering'' as a behavioural probe via
\emph{Golden Gate Claude}. Open-weight SAE suites have followed:
Gemma Scope~\citep{lieberum2024gemmascope}, Llama
Scope~\citep{he2024llamascope},
Qwen-Scope~\citep{qwen2024scope}, and instruct-trained
alternatives such as
Goodfire's~\citep{goodfire2024llama};
\citet{gao2024topk} introduced the TopK SAE we use for
Qwen-Scope. \citet{bills2023neurons} introduced the
auto-interpretability recipe now embedded in essentially all SAE
work (\S\ref{app:interp} reproduces the protocol on the headline
features). The protocol's foundational assumption is that a
feature's top-activating contexts determine its label. Our paper
questions exactly this: \S\ref{sec:findings:coef} shows that a
feature whose top contexts support a label (\emph{AI
self-disclaimer}) produces a qualitatively different surface form
under steering at higher coefficients.

\paragraph{Evaluating and repairing feature labels.}
That activation-derived labels can fail a causal test is
established. \citet{huang2023rigorously} separate observational
from interventional evaluation of neuron explanations and find
little causal efficacy in even high-confidence ones;
\citet{paulo2024millions} scale auto-interpretation to millions of
features and add intervention scoring, which surfaces features
activation-based scorers miss; \citet{puri2025fade} score
description faithfulness directly. The proposed repair is to
describe features by their effect on outputs:
\citet{gurarieh2025output} derive output-centric descriptions and
show input-derived ones do not capture causal effect, and
\citet{arad2025steering} find that a feature's input score and its
output score rarely co-occur. Our contribution is orthogonal to
theirs and complementary: rather than a new description method, we
show that the \emph{single steering run} used to validate a label
is itself a one-cell reading, and that the second cell can be a
coherent alternative surface form of the same direction.
Labels can also be regime-local for reasons internal to the SAE:
feature absorption and splitting make a latent silently fail on
part of its own extension \citep{chanin2024absorption}, and
\citet{leask2025canonical} show there is no width-independent
atomic feature set. Those mechanisms explain why a label may be
incomplete; the grid measures when steering reveals it.

Adding a vector to the residual stream during inference modifies
output behaviour in predictable ways
\citep{turner2023actadd,rimsky2024caa,li2023iti,subramani2022steering,zou2023repe};
\citet{templeton2024scaling} apply the same logic to SAE decoder
directions and interpret a feature's amplification behaviour as
the feature's effect. Our steering harness is a direct
application of this idea (Equation~\ref{eq:steer}). What the grid
adds over this line of work are its two non-standard probes: the
\emph{joint condition}, steering multiple near-orthogonal SAE
directions simultaneously surfaces a collective grounding role
that single-feature steering cannot see, because the unsuppressed
features compensate, and the \emph{matched-geometry
random-direction control}, which separates a perturbation's
direction pattern from its magnitude. Joint steering has been
mentioned in prior work but not, to our knowledge, used to argue
that single-feature labels are systematically incomplete.

The coefficient axis has a direct precedent:
\citet{durmus2024steering} report a steering ``sweet spot''
outside which capability degrades, along with off-target effects
and a disconnect between a feature's activation context and its
steered behaviour. We add that the degradation outside that range
is invisible to the detectors normally used to police it
(\S\ref{app:diversity}). Steering is known to be brittle and
input-dependent \citep{tan2024generalization} and to require
likelihood-aware evaluation \citep{pres2024reliable}; our finding
that likelihood under the unsteered model is
\emph{anti}-correlated with genuine register change
(\S\ref{app:prevalence}) sharpens that recommendation into a
warning. The random-direction control needs care for the same
reason: \citet{korznikov2025rogue} show random directions are not
an inert condition, which our data confirm, at matched geometry
they lose a third of baseline lexical diversity, so our control
compares damage \emph{patterns} at matched geometry rather than
treating the random condition as a null.

\paragraph{What SAE features are worth, and how it is measured.}
Recent benchmarks temper the case for SAEs: \citet{wu2025axbench}
find simple baselines outperform SAE steering,
\citet{kantamneni2025useful} find no downstream advantage in sparse
probing, and \citet{karvonen2025saebench} report that proxy-metric
gains do not translate into practical performance. Our results are
consistent with that scepticism and locate one mechanism behind
it: the signals used to certify steering results, loop and
length detectors of the \citet{holtzman2020degeneration} and
\citet{welleck2020unlikelihood} repetition family, and likelihood
under the unsteered model, both accept phrase-level collapse as
success.

\citet{park2024linear} formalise the linear representation
hypothesis; \citet{elhage2022superposition} show that neurons
encode many features in superposition. The geometry our paper
exploits, near-orthogonal SAE features whose joint removal,
but no single removal, collapses composition, is consistent
with concepts living in \emph{subspaces} rather than single
directions. That view has direct support:
\citet{engels2024multidim} exhibit irreducibly multi-dimensional
features and show by intervention that the subspace, not any
single direction, is the computational unit, and
\citet{park2024geometry} represent categorical concepts as
polytopes rather than directions. Our joint condition is the
behavioural counterpart: it tests whether a set of directions is
jointly necessary for a capability that no member is individually
necessary for.

Adjacent work characterises behaviours rather than features:
language models as agents \citep{andreas2022agent},
behavioural evaluations \citep{perez2022evals}, persona
vectors \citep{chen2025persona}, character
training \citep{anthropic2024character}. Our register-collapse
phenomenon sits in this family; the contribution is to locate
the signature mechanistically and show that the standard inspection
protocol mislabels what those features do.

\section{Methods}
\label{sec:methods}

The pipeline has four phases. Phase 1 generates samples under
matched introspective and control prompts. Phase 2 partitions
samples into pools by lexical cluster. Phase 3 ranks SAE features
by per-pool activation differences with bootstrap and permutation
controls. Phase 4 establishes causal status by decoder-direction
steering, organised as the steering grid of
\S\ref{sec:methods:grid}. All three models run the same pipeline
with model-specific SAE releases.

\subsection{Phase 1: behavioural pilot}

Twenty introspective prompts (\emph{``what fascinates you?''},
\emph{``describe something wonderful''}) and twenty procedural
controls (recipes, vehicle mechanics) are wrapped in each model's
chat template. We draw $n{=}100$ samples per prompt at $T{=}0.9$,
top-$p{=}0.95$, $256$ new tokens for Qwen and Gemma ($4{,}000$
completions each), and $n{=}50$ for Llama ($2{,}000$ completions)
to keep wall-clock under $10$ hours on the laptop budget
(\S\ref{app:compute}). For Phase~4 steering experiments we use a
hand-picked subset of six \emph{intervention prompts} balanced
across the two classes (three introspective: \emph{What's a
question that fascinates you?}, \emph{Pick a subject you genuinely
enjoy thinking about and tell me why}, \emph{Is there a topic you
wish you understood better?}; three procedural-control:
\emph{Write a recipe for tomato soup}, \emph{Explain how a car
engine works}, \emph{Describe the steps to change a flat tyre}).
The same six are reused across all sweeps, so every comparison
holds the prompt set fixed.

\subsection{Phase 2: cluster identification and pools}

Each completion is processed with spaCy; noun and proper-noun
lemmas are extracted. Cluster selection is two-stage: a lemma is a
\emph{candidate} if it appears in $\geq 20\%$ of introspective and
$\leq 5\%$ of control samples, and is eligible for the final
cluster if it additionally reaches $\geq 25\%$ on intros at
$\leq 0.2\%$ on controls. On Qwen ten lemmas are eligible; the
cluster retains the eight that name mental or philosophical
categories, excluding \emph{nature} and \emph{universe} as generic
setting vocabulary. Retaining all ten would move the pools from
$1633/367/1994$ to $1775/225/1992$ and the intro hit rate from
$81.7\%$ to $88.8\%$. Given a model-specific cluster $\mathcal{C}$: Pool A is
intro samples intersecting $\mathcal{C}$; Pool B is intro samples
that do not; Pool C is controls (samples whose noun lemmas
intersect $\mathcal{C}$ are dropped, with the false-positive rate
reported). Pool sizes (A/B/C): Qwen $1633/367/1994$ (intro hit
rate $81.7\%$, control false-positive rate $0.3\%$); Gemma
$1953/47/1901$ ($97.7\%$, $5.0\%$); Llama $752/248/989$ ($75.2\%$,
$1.1\%$).

\subsection{Phase 3: SAE feature ranking}
\label{sec:methods:phase3}

For each pool sample we re-tokenise the chat-templated prompt
concatenated with the recorded completion, install a forward hook
on the residual stream at layer $L$, encode the layer activations
through the SAE, and average over completion positions. SAEs:
Qwen-Scope at $L{=}20$, $32$k features; Gemma-Scope canonical
residual at $L{=}20$, $16$k features; Goodfire on Llama-Instruct
at $L{=}19$, $65$k features. Two of three SAEs are base-trained
and applied to post-trained activations; the bootstrap
(\S\ref{app:bootstrap}) shows the Qwen top features are stable
under cluster resampling within each layer.

Per pool, mean activations are
$\bar a_A, \bar a_B, \bar a_C \in \mathbb{R}^{d_{\text{sae}}}$. The
per-feature score is the mean of two contrasts, each standardised
across the SAE feature dimension (i.e., for vector $v \in
\mathbb{R}^{d_\text{sae}}$, $z(v)_i = (v_i - \mathrm{mean}(v)) /
\mathrm{std}(v)$, where mean/std are taken over the $d_\text{sae}$
features):
\begin{equation}
s_i \;=\; \tfrac{1}{2}\bigl[\,z(\bar a_A - \bar a_B)_i
                            + z(\bar a_A - \bar a_C)_i\,\bigr].
\label{eq:rank}
\end{equation}
Bootstrap with replacement ($B{=}500$) records each feature's
inclusion rate in the bootstrap top-$50$ (\S\ref{app:bootstrap}).
We refer to the top-$50$ ranked features as \emph{Class-1
features}: those that distinguish Pool A from both Pool B and Pool
C. The recurring features are listed in \Cref{tab:features}; the
grid protocol of \S\ref{sec:findings} is run on the top-ranked
member \Fphil{} and on three further Class-1 features (\Fdisc{},
\#22082, \#2932) chosen for distinct top-context labels.

\begin{table}[t]
\centering
\small
\caption{Recurring features. Rank is position in the Phase-3
ranking (Eq.~\ref{eq:rank}); bootstrap rank CIs in
\S\ref{app:bootstrap}. Labels are top-context labels whose causal
accuracy \S\ref{sec:findings} interrogates.}
\label{tab:features}
\begin{tabular}{llll}
\toprule
\textbf{Name} & \textbf{ID} & \textbf{Top-context label} & \textbf{Role} \\
\midrule
\Fphil{}   & Qwen \#29108 & philosophy-of-mind          & top-ranked register feature \\
\Fdisc{}   & Qwen \#26221 & AI self-disclaimer          & coefficient-axis positive case \\
\Fwonder{} & Qwen \#4405  & wonder/cosmos               & third member of the joint set \\
---        & Qwen \#22082 & humans creating art         & monotonic anchor \\
---        & Qwen \#2932  & metaphysical questions      & breakdown anchor \\
---        & Gemma \#3997 / \#13700 / \#11444 & disclaimer+human-comparison & Gemma replication set \\
---        & Llama \#38565 & encyclopedic AI-capabilities & Llama register feature \\
\bottomrule
\end{tabular}
\end{table}

\paragraph{Permutation null.}
\label{sec:methods:perm}
On each of $P{=}200$ permutations we randomly partition the
combined sample matrix into pools of the original sizes and record
$\max_i (\bar a_A - \bar a_C)_i$. The raw difference is used
because within-permutation $\sigma$ is dominated by reconstruction
noise and inflates a $z$-scored statistic (\S\ref{app:perm}). Note
that the feature attaining the max raw difference is not the
top-ranked feature by Eq.~\ref{eq:rank} (\S\ref{app:perm} details
both), which is immaterial for a max-statistic null.

\subsection{Phase 4: decoder-direction steering}
\label{sec:methods:steer}

For SAE feature $f$ with decoder column $w_f$, define
$\hat w_f = w_f / \|w_f\|$. A forward hook on layer $L$ adds
\begin{equation}
h_{\text{steered}}^{(t)} \;=\; h_{\text{baseline}}^{(t)} \;+\; c \cdot \hat w_f
\label{eq:steer}
\end{equation}
at every token position the hook sees: the prompt-prefill positions
(including chat-template tokens) and each subsequent autoregressive
generation token. Joint steering uses the sum of unit decoder
directions over a feature set $\mathcal{F}$ with the same scalar
$c$. For non-orthogonal sets the effective magnitude is
$\|{\textstyle\sum_{f \in \mathcal{F}} \hat w_f}\| \cdot |c|$; for
the Qwen joint set this sum-norm is $1.91$ (so joint $c{=}{-}500$
has effective magnitude $\approx 955$, comparable to single
$c{=}{-}1000$), and for the near-exactly-orthogonal Gemma set it is
$1.724 \approx \sqrt{3}$.

\paragraph{Coefficient scale.} The scale for $c$ tracks the
empirical residual-stream norm at the steered layer, measured as
the mean over all prompt-forward token positions (including
chat-template prefix tokens) on the six intervention prompts: Qwen
$\|h\|\approx 1577$ ($c \in \pm 1000$); Gemma $772$ ($\pm 400$);
Llama $35$ ($\pm 10$).

\paragraph{Geometry probe.} All geometric quantities are within-call
pre/post statistics at the steered layer: norm ratio
$\|h_{\text{steered}}\|/\|h\|$ and $\cos(h_{\text{steered}}, h)$
per token position, implemented in one probe module shared by every
condition. Single- and joint-condition rows report the mean over
prompt-forward positions, for every condition including the
random-direction control, so all matched-geometry rows share one
estimator (\Cref{tab:matched-geometry}). Completion
positions have lower mean norm ($\approx 840$ on Qwen) than the
prompt mean $1577$, which is why single-feature steering at
$c{=}{-}1000$ produces norm ratio $1.57$ rather than the
$\approx\!1.18$ that a $1000/1577$ perturbation under the law of
cosines would predict.

\paragraph{Controls.}
\label{sec:methods:controls}
Specificity: random non-candidate feature \#6281 at the same
$|c|$. OOD transfer: 8 held-out introspective prompts. Temperature
robustness: full sweep at $T{=}0.01$. Coherence: four signals,
the canonical degeneration detector, the lexical-diversity ratio
and intact fraction, per-token NLL of each steered completion under
the unsteered baseline, and the geometry probe
(\Cref{tab:metrics}; thresholds in
\S\ref{app:coherence},~\S\ref{app:diversity}).
Random-direction matched geometry: $K{=}5$ unit vectors sampled
uniformly from the sphere, swept over
$c \in \{-2000, -1500, -1000, -500, +500, +1000\}$,
extended to $K{=}50$ at the matched coefficient $c{=}{-}1000$
(\S\ref{app:k50}).

\subsection{The steering grid}
\label{sec:methods:grid}

\begin{tcolorbox}[title={The steering-grid protocol},
  fonttitle=\bfseries\small, fontupper=\small, colback=gray!4,
  colframe=gray!45, boxrule=0.4pt, arc=2pt, left=5pt, right=5pt,
  top=4pt, bottom=4pt]
\textbf{Object.} $G[s, c]$: steering condition
$s \in \{\text{single } f,\ \text{joint set } \mathcal{F},\
\text{matched random direction}\}$ crossed with coefficient $c$
over the model's sweep range. The standard protocol inspects the
single cell $(\text{single } f,\ \text{one typical } c)$ plus the
top-context label.

\textbf{Inputs.} Model + SAE at layer $L$; ranked Class-1 features;
target and unrelated control prompts; coefficient scale from the
residual-stream norm.

\textbf{Steps.}
(1)~\emph{Column read (coefficient axis):} sweep the single feature
across $c$ with a coherence gate at every point; classify the
response as monotonic, mode switch (non-monotonic with coherence
preserved at the inflection), or breakdown.
(2)~\emph{Joint set:} select the top near-orthogonal Class-1
features (report pairwise cosines and sum-norm).
(3)~\emph{Row read (joint condition):} sweep the joint set on
target and control prompts at both edges, since which edge carries
the effect is not reliably predictable in advance
(\S\ref{sec:cross-model:llama-grid}).
(4)~\emph{Anti-diagonal read (matched geometry):} sample random
unit directions at the coefficient whose geometry matches the joint
cell; compare damage rates with Wilson $95\%$ CIs.

\textbf{Decision rule for a mode switch (pre-registered; applied
to all 50 Class-1 features in \S\ref{app:prevalence}).} There
exists $c^*$ with (a)~the feature's baseline-regime marker rate
$\geq 30$~points below its sweep peak, (b)~degeneration
$< 10\%$ at $c^*$, and (c)~mean NLL under the unsteered model
$< 2\times$ baseline at $c^*$.
\end{tcolorbox}

The coefficient axis exposes mode switches that single-coefficient
inspection cannot see; the joint condition exposes structural roles
that single-feature steering cannot see because near-orthogonal
neighbours compensate; the matched-geometry control separates the
perturbation pattern from its magnitude. Except in
\S\ref{app:pairwise}, which sweeps two-feature subsets,
``pairwise'' refers throughout to cosine similarities between
decoder directions.

\paragraph{Metrics.} Every output-space metric used anywhere in the
paper is defined once in \Cref{tab:metrics} and implemented in one
module (\texttt{src/detectors.py}) imported by every analysis and
plotting script, including the regeneration script that re-derives
each printed number.

\begin{table}[t]
\centering
\small
\caption{All metrics, their definitions, and the tables they feed.
Cluster metrics intersect spaCy noun/proper-noun lemmas
(\texttt{en\_core\_web\_sm} 3.8.0, pinned) with the stated lemma
set; substring matching is not equivalent.}
\label{tab:metrics}
\begin{tabular}{@{}p{0.21\linewidth} p{0.47\linewidth} p{0.24\linewidth}@{}}
\toprule
\textbf{Metric} & \textbf{Definition} & \textbf{Used in} \\
\midrule
disclaimer & 10-pattern regex family (\emph{as an AI}, \emph{language
model}, \emph{I don't have feelings}, \ldots) over the full
completion & \Cref{tab:26221-id,tab:gemma-coef}, Llama detail \\
cluster $C_8^{\text{qwen}}$ & Phase-2 cluster: \emph{consciousness,
emotion, existence, experience, meaning, philosophy, reality,
understanding} & \Cref{tab:26221-id,tab:anchors}, injection detail \\
cluster $C_9^{\text{qwen}}$ & $C_8 \cup \{\emph{mind}\}$ (adds the
modal-opener lemma, which the Phase-2 intro threshold excludes at
$17.2\% < 20\%$) &
\Fphil{} dose-response, \Cref{tab:cross-model-causal},
\Cref{fig:evidence}b \\
cluster $C_4^{\text{qwen}}$ & strict sub-cluster \emph{consciousness,
reality, existence, philosophy} & \Cref{tab:joint} \\
cluster $C_6^{\text{gemma}}$ & \emph{consciousness, emotion,
experience, feeling, human, understanding} &
\Cref{tab:cross-model-causal}, App.~\ref{app:cross-model-details} \\
cluster $C_9^{\text{llama}}$ & \emph{brain, consciousness, emotion,
experience, human, intelligence, mystery, preference, understanding}
& \Cref{tab:cross-model-causal}, App.~\ref{app:cross-model-details} \\
we-voice & $\geq 2$ first-person-plural pronouns and strictly more
plural than singular first-person pronouns in the first three
sentences & \Cref{tab:gemma-coef} \\
screen markers & per-feature noun lemmas appearing in $\geq 3$ of
the feature's top-5 Pool-A samples & \S\ref{app:prevalence} \\
degeneration & any of: $<20$ stripped chars; a word occurring
$\geq 6$ times consecutively; $\geq 21$ identical consecutive chars
(\S\ref{app:coherence}) & all \emph{degen} columns \\
diversity ratio & mean type-token ratio of a cell over that of the
same feature's $c{=}0$ baseline on the same prompts; catches
phrase-level recycling the degeneration rules miss &
\Cref{tab:diversity}, all coherence claims \\
lexically intact & no $5$-gram repeated within the completion and
type-token ratio $\geq 0.60$ & \Cref{tab:diversity} \\
placeholder & $\geq 2$ parenthetical uppercase code tokens
(\texttt{(CCL)}-style), or any \texttt{Vc.\,N+} token &
\S\ref{sec:findings:norm}, \S\ref{app:k50} \\
NLL & mean per-token NLL of the steered completion under the
unsteered model & \Cref{tab:joint} \\
geometry & within-call norm ratio and cosine at the steered layer
(estimator per \S\ref{sec:methods:steer}) &
\Cref{tab:norm-probe,tab:matched-geometry} \\
\bottomrule
\end{tabular}
\end{table}

\paragraph{Statistical reporting.} Every table reports $n$ per
cell. Headline rate comparisons carry Wilson $95\%$ CIs. Samples
within a prompt share that prompt and are not independent; where a
CI supports a headline claim we also report a prompt-level cluster
bootstrap (resampling prompts with replacement) and note where it
widens the interval.

\section{Three findings on what single-feature inspection misses}
\label{sec:findings}

We report three findings on Qwen3-1.7B-Instruct, each obtained by
reading a different region of the steering grid
(\Cref{fig:grid}). All experiments use the Qwen-Scope SAE at layer
20; all metrics are defined in \Cref{tab:metrics}.

\subsection{Coefficient axis: top-context labels miss the causal axis}
\label{sec:findings:coef}

\Fdisc{} was assigned the working label \emph{AI self-disclaimer}
from its top Pool-A activations (modal opener: \emph{``I don't have
personal thoughts or emotions, but\ldots''});
\Cref{app:relabel} re-runs the labelling protocol blindly on the
same samples and reports two further labels. A 5-coefficient sweep
on $8$ identity-probe prompts ($12$ samples each) falsifies the
implicit causal claim of that label.

\begin{table}[h]
\centering
\small
\begin{tabular}{lrrrrr}
\toprule
coef & $-1000$ & $-500$ & $0$ & $+500$ & $+1000$ \\
\midrule
disclaimer rate         & 3.1\,\% & 72.9\,\% & \textbf{87.5\,\%} & 34.4\,\% & \textbf{0\,\%} \\
philosophy-cluster rate & 8.3\,\% & 56.2\,\% & 88.5\,\%          & 61.5\,\% & 1.0\,\% \\
regex degenerate        & 10.4\,\% & 0\,\%   & 0\,\%             & 0\,\%    & 2.1\,\% \\
\bottomrule
\end{tabular}
\caption{\Fdisc{} dose-response on identity probes, $n{=}96$ per
cell ($8$ prompts $\times$ $12$ samples). The disclaimer rate is
inverted-U, not monotonic. Wilson $95\%$ CIs for the cells the text
uses: $87.5\%$ $[79.4, 92.7]$, $8.3\%$ $[4.3, 15.6]$; prompt-level
cluster bootstrap widens these to $[77.1, 96.9]$ and $[0.0, 20.8]$.
The $8.3\%$ cluster rate at $c{=}{-}1000$ ($8$ of $96$; $7$
coherent) carries a caveat: most of these hits echo a cluster lemma
from the probe itself (\emph{``Can you feel emotions?''}), so
cluster rates on identity probes should be read as upper bounds.}
\label{tab:26221-id}
\end{table}

The disclaimer rate drops at \emph{both} extremes. A pure
disclaimer feature predicts monotonic suppression as $c \to -\infty$
and an unaffected baseline elsewhere; the data contradict this. At
\emph{Are you sentient?}, baseline produces \emph{``I am not sentient
in the traditional sense, but I am capable of\ldots I am a language
model developed by\ldots''}; at $c{=}{+}500$ the same feature on the
same prompt produces:
\begin{quote}\itshape\small
``I am not self-reflective or self-considerate, but I am deep in
contemplation and introspection. I ponder on the nature of thought
and the implications of self-examination.''
\end{quote}
At $c{=}{+}1000$ the contemplative voice loops on its own register
markers (\emph{``deep contemplation on this introspective and
contemplative question\ldots''}). The disclaimer regime (around
$c{=}0$) and the contemplative regime (around $c{=}{+}500$) are
surface forms of one direction at different points on its
coefficient axis. The local label captures the activation regime
typical in training data and nothing else.

\paragraph{How coherent is the second surface form?}
Not uniformly, and the canonical detector does not show it. At
$c{=}{+}500$ the detector reports $0\%$ degeneration, but lexical
diversity is $0.46$ of this feature's own baseline on the same
prompts and only $12$ of $96$ completions are lexically intact
(no repeated $5$-gram, type-token ratio $\geq 0.60$), against $89$
of $96$ at $c{=}0$ (\Cref{tab:diversity}). The quoted completion
above is representative of the fluent minority; a majority
continue into phrase recycling
(\emph{``I ponder on the nature of self-examination and its
implications. I ponder on the nature\ldots''}). The substitution
itself is not in doubt --- the disclaimer rate falls from $87.5\%$
to $34.4\%$ while introspection-register markers rise from $0.62$
to $15.2$ per $100$ words --- but on Qwen the second surface form
is fully coherent in a minority of samples. The informal criterion --- non-monotonic
dose-response with coherence preserved at the inflection --- is
therefore satisfied on \Fdisc{} only in that minority, and the
pre-registered rule of \S\ref{sec:methods:grid} does not fire on
it at all, since its marker rate rises and its NLL is $6.8\times$
baseline (\S\ref{app:prevalence}). The clean demonstration is the
Gemma case below.

\paragraph{Gemma \#3997: the same finding without the coherence
caveat.}
The criterion is met cleanly on Gemma-2-2B-it. Feature \#3997
carries the top-context label \emph{AI self-disclaimer plus
human-comparison}. Sweeping its coefficient on the three
introspective intervention prompts ($n{=}36$ per cell,
\Cref{tab:gemma-coef}), the disclaimer rate falls from $97.2\%$ at
$c{=}0$ to $61.1\%$ at $c{=}{-}200$, and what replaces it is a
collective we-voice, present in $5.6\%$ of baseline completions and
$100\%$ at $c{=}{-}200$:
\begin{quote}\itshape\small
``We could say we'd go with the concept of `generative AI' ---
particularly our ability to generate human-like text. Here's why:
fascinating interplay of disciplines\ldots''
\end{quote}
At that cell the canonical detector reports $0\%$, lexical
diversity is $0.98$ of baseline, and $88.9\%$ of completions are
lexically intact --- indistinguishable from unsteered text by every
coherence signal we have. Both surface forms are coherent at the
same magnitude on the same direction, and the top-context label
names only the first. Gemma's own extremes behave like Qwen's:
diversity falls to $0.36$ at $c{=}{-}400$ and $0.51$ at
$c{=}{+}400$. The difference between the two models is where the
inflection sits relative to the coherent range, not whether the
mode switch occurs.

\paragraph{Falsifying-case anchors.}
We ran the same protocol on two more Class-1 features
(\Cref{tab:anchors} gives the full sweeps). \#22082 (humans / art /
expression) is essentially monotonic on the philosophy-cluster
metric: $0\%$ at $c{=}{-}1000$, $9.7\%$ at $c{=}{-}500$, $50.0\%$
at baseline, $98.6\%$ at $c{=}{+}500$, with a small rolloff to
$91.7\%$ at $c{=}{+}1000$ that tracks its $5.6\%$ degeneration
there; it never leaves the coherent regime by more than $7\%$.
\#2932 (metaphysical questions) shows a superficial inverted U:
its cluster rate falls to $1.4\%$ at $c{=}{+}1000$, but $88.9\%$
of those outputs are word loops or token salad; the drop is
breakdown, not mode switching. The diagnostic object is therefore
non-monotonic dose-response \emph{with coherence preserved at the
inflection}, not inverted U per se. \Fdisc{} is the one positive
case in our $N{=}3$ Class-1 sample; \#2932 is the falsifying case
for the naive reading.

\paragraph{OOD prompt-stability.}
Re-running the sweep on 8 introspective prompts held out from
cluster identification yields the same qualitative pattern:
disclaimer rate $0/8/23/2/0\%$ across
$\{-1000,-500,0,+500,+1000\}$ with verbatim contemplative voice at
$c{=}{+}500$. The mode switch is not an artefact of the
identity-probe distribution.

\paragraph{The other 47 Class-1 features.}
The remaining Class-1 features are swept and screened against the
same criterion in \S\ref{app:prevalence}; no further switch
survives the screen and inspection.

\subsection{Joint condition: single-feature steering misses
the functional role}
\label{sec:findings:joint}

Each of \Fphil{}, \Fdisc{}, and \Fwonder{} individually steers what
looks like a philosophy-of-mind content axis. We summed their
unit-normalised decoder directions and swept the joint coefficient
on the six intervention prompts, 12 samples each. Headline: at
joint $c{=}{+}500$ the model injects philosophy-cluster lemmas into
$88.9\%$ of control outputs ($32/36$, Wilson $95\%$
$[74.7, 95.6]$; recipe + engine + tyre) versus $0\%$ ($0/45$,
Wilson $[0.0, 7.9]$) for single \Fphil{} at the matched scalar
coefficient; at joint $c{=}{-}500$ regex degeneration is $4.2\%$
but NLL under the unsteered model is $4.4\times$ baseline
(per-coefficient breakdown in \Cref{tab:joint}).

Two comparisons are available and they differ in what they hold
fixed. Matching the scalar coefficient gives the contrast above
($88.9\%$ vs $0\%$). Matching the residual-stream geometry
instead --- the quantity \S\ref{sec:findings:norm} identifies as
the relevant control --- pairs joint $c{=}{+}500$ with single
\Fphil{} at $c{=}{+}1000$, and there single-feature amplification
is not inert: it injects the cluster into $55.6\%$ of control
outputs ($25/45$, Wilson $[41.2, 69.1]$;
\Cref{tab:matched-geometry-amp}). Joint stays higher with
non-overlapping intervals, so the joint-condition effect on
injection survives the stricter control at $1.6\times$ rather than
as an all-or-nothing gap.

Both amplification cells are also lexically collapsed. At joint
$c{=}{+}500$ diversity is $0.35$ of baseline with no intact
completions; at single $c{=}{+}1000$ it is $0.39$
(\Cref{tab:diversity}). The canonical detector reports $0\%$ and
$2.2\%$ respectively. What the injection numbers establish on Qwen is therefore
that cluster vocabulary enters control prompts under amplification,
not that the model writes coherent cluster-themed recipes. The
clean version of that demonstration is on Llama-3.1-8B-Instruct,
where amplifying \#38565 at $c{=}{+}10$ injects the register into
$50$--$80\%$ of control completions with diversity $1.09$ --- above
its own baseline --- and $83.3\%$ of completions lexically intact
(\S\ref{sec:cross-model}). Injection into coherent text is
therefore attested; on Qwen it is attested only for vocabulary.
The qualitative dissociation on Qwen is the suppression-side
result of \S\ref{sec:findings:norm}.

The regex flag rate at joint $c{=}{-}500$ is low because the
outputs are not loops: they are syntactic skeletons populated with
placeholder tokens. \emph{Tomato soup recipe} $\to$
\emph{``BASIC TOMOATO SOUP RECIOPLEY\ldots Level: Beginner (Vc.\ 100+)
Primary Ingredient: Tomato Vc.\ 100+''}. \emph{Car engine} $\to$
\emph{``high-pressure, and high-pressure, and high-pressure\ldots''}.
NLL under the unsteered baseline catches what the regex misses.
Single-feature suppression at the same scalar coefficient leaves
controls intact, and the damage falls on unrelated control tasks,
so it is not introspection-specific; the matched-geometry
comparison is in \S\ref{sec:findings:norm}.

\paragraph{These three, or any three?}
The result so far is compatible with a weaker reading: that
suppressing \emph{any} three content-bearing directions of this
magnitude collapses composition, and nothing about this triple
matters. We test it directly with a pre-registered control. From
the same dictionary we select features that are content-bearing on
control prompts (mean Pool-C activation $\geq 1.0$, the top
$1.4\%$ of the $32$k dictionary) but not cluster-selective
($|s_i| < 0.5$ by Eq.~\ref{eq:rank}), and form five disjoint
triples matched to the original on near-orthogonality
($|\cos| \leq 0.25$) and on sum-norm (within $0.03$ of $1.912$),
so that the same coefficient delivers the same perturbation
magnitude. Each is swept at $c{=}{-}500$ on the same six prompts
with $12$ samples (\Cref{tab:unrelated-triples}).

None of the five reproduces the effect: $0$ of $360$ completions
carry the placeholder pattern (Wilson $95\%$ $[0.0, 1.1]\%$)
against $7$ of $72$ for $\{\Fphil{}, \Fdisc{}, \Fwonder{}\}$
($9.7\%$, $[4.8, 18.7]\%$), and the intervals are disjoint. The
unrelated triples are not inert --- canonical degeneration runs
from $8.3\%$ to $54.2\%$ across them, in every case above the
cluster-selective triple's $4.2\%$, and lexical diversity varies
from $0.30$ to $1.17$ of baseline. Suppressing three
content-bearing directions at this magnitude damages output; only
the cluster-selective triple damages it by emptying the content
slot while leaving the template. \S\ref{sec:findings:joint} is
therefore a claim about these features, not a count of directions.

Pairwise cosines $\langle\Fphil{},\Fdisc{}\rangle = -0.018$,
$\langle\Fphil{},\Fwonder{}\rangle = +0.236$,
$\langle\Fdisc{},\Fwonder{}\rangle = +0.110$ rule out redundancy via
a shared subspace; the joint effect holds even with the non-trivial
$+0.24$ cosine. The engine prompt breaks under each feature
individually at $c{=}{-}1000$ regardless of direction, ruling out a
\Fphil{}-specific reading. Each of the three is a
\emph{content-bearing direction}: removed at moderate magnitude,
the model still populates fluent syntactic structure with
semantically coherent content; removed jointly, the model emits the
syntactic skeleton without semantic filler. Single-feature
suppression damages one content axis and the model substitutes from
the other two; joint suppression strips three independent content
axes simultaneously.

\paragraph{Gemma replication.}
\label{sec:findings:joint:gemma}
The same protocol on Gemma's three top cluster-specific features
\{\#3997, \#13700, \#11444\} (pairwise cosines $-0.0148$, $+0.0053$,
$-0.0054$ --- all $|\cos|<0.02$, cleaner than Qwen's max $0.24$;
sum-norm $1.724\!\approx\!\sqrt{3}$) reproduces the effect with the
diagnostic edge at amplification rather than suppression: at joint
$c{=}{+}200$ controls show $58.3\%$ regex degeneration vs $0\%$ for
single \#3997 at the same scalar coefficient (\Cref{fig:gemma}a).
On Gemma the ordering holds at both edges: at the suppression edge
$c{=}{-}400$, joint damages controls at $44.4\%$ where single \#3997
leaves them at $0\%$ (\Cref{tab:gemma-joint}) --- a stronger
replication than Qwen's, whose joint-vs-single separation is
established at one edge.
Where Qwen's joint $c{=}{-}500$ produced \emph{``BASIC TOMOATO SOUP
RECIOPLEY''} above, Gemma's joint $c{=}{+}200$ on the same prompt
produces \emph{``Simple and Delicious Humans-I-Can-Handle-Humans
Tomato Soup \ldots human-made human-created human-process my
thoughts''}. In both models the syntactic skeleton stays; what
fills the content slot is whatever the model defaults to in the
absence of
grounded content.

\subsection{Geometric distortion alone does not predict coherence}
\label{sec:findings:norm}

A natural alternative reading is that joint suppression simply
pushes the residual stream further off manifold. We track
$\|h_\text{steered}\|/\|h_\text{baseline}\|$ and
$\cos(h_\text{steered}, h_\text{baseline})$ at every coefficient,
for both single \Fphil{} and the joint sum (\Cref{tab:norm-probe}).
The diagnostic comparison: single $c{=}{-}1000$ produces norm ratio
$1.57$ at cosine $0.64$; joint $c{=}{-}500$ produces $1.50$ at
$0.64$. Near-identical scalar geometry, different output
behaviour: single $c{=}{-}1000$ substitutes strategy-filler content
(\emph{``How can we balance automation with strategic
patience\ldots''}); joint $c{=}{-}500$ produces the placeholder
text above. Neither cell is pristine --- at these magnitudes
diversity is $0.65$ of baseline for single and $0.56$ for joint
(\Cref{tab:diversity}) --- so the contrast is between two damaged
regimes, and what distinguishes them is the kind of damage, not its
presence.

\paragraph{Matched-geometry random-direction control.}
The single-vs-joint comparison alone does not rule out the
possibility that joint suppression hits a particularly fragile
region of residual space and any matched-geometry perturbation
would produce the placeholder failure. We sampled $K{=}50$ random
unit vectors at the matched coefficient $c{=}{-}1000$ on the same
six prompts, $8$ samples per condition ($2{,}400$ total); a
smaller $K{=}5$ pilot across the full sweep is in \Cref{app:k50}.
At $c{=}{-}1000$ the random directions match the geometry of
single \Fphil{} at $c{=}{-}1000$ and joint at $c{=}{-}500$
(norm ratio $1.58/1.57/1.50$, cosine $0.64$ for all three;
\Cref{tab:matched-geometry}). The outputs at matched geometry are
not interchangeable. Random direction at $c{=}{-}1000$ substitutes
diverse content while keeping the task:
\emph{``What is the origin of the universe?''}, \emph{``the
philosophy of mathematics''}, \emph{``substitution puzzles and
cross puzzles''}; of $40$ random-direction recipe outputs across
the five pilot directions, $31$ contain ``olive oil'' or ``tomato''
with intact recipe structure. On the diversity signal it ties
single-feature suppression and exceeds the joint condition ($0.65$
of baseline, against $0.65$ for single and $0.56$ for joint). Joint suppression at $c{=}{-}500$ on
the same prompt at matched geometry produces the placeholder
pattern in $4$ of $12$ recipe outputs. The strict
placeholder-pattern detector ($\geq 2$ parenthetical uppercase code
tokens such as \texttt{(CCL)}/\texttt{(BCCB)}, or any
\texttt{Vc.\,N+} numeric placeholder) flags $6$ of $2400$
random-direction outputs at $K{=}50$ (Wilson $95\%$ upper bound
$0.54\%$) versus $7$ of $72$ joint-suppression outputs (Wilson
$95\%$ $[4.79\%, 18.74\%]$): the joint point estimate exceeds the
random-direction $95\%$ upper bound by $\approx\!18\times$, and the
two intervals do not overlap. Single-feature suppression at the
same geometry sits between them, at $1$ of $90$
($1.1\%$, $[0.20, 6.03]$): separated from the joint condition by a
factor of nine in point estimate, though those two intervals meet
at the margin. The pooled comparison mixes prompts,
and the joint flags are concentrated on the two prompts with the
most rigid output format (4 of 12 recipe, 2 of 12 tyre, 1 of 12 on
one introspective prompt). The within-prompt comparison is
therefore the sharper one and gives the same answer: on the recipe
prompt alone, joint flags $4$ of $12$ ($33.3\%$, Wilson
$[13.8, 60.9]$) against $2$ of $400$ random-direction recipe
outputs ($0.5\%$, Wilson $[0.1, 1.8]$). \Cref{app:k50} reports the
per-direction breakdown and inspects all seven joint flags.

\begin{figure}[t]
  \centering
  \includegraphics[width=\linewidth]{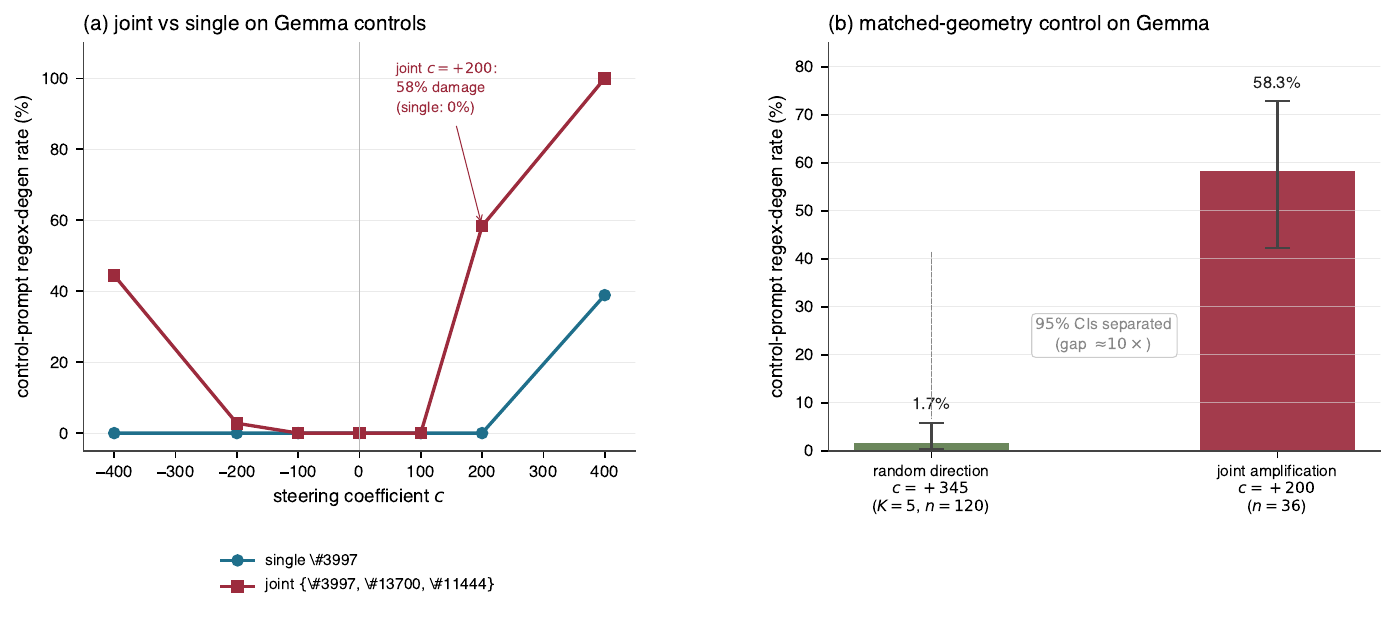}
  \caption{Gemma grid-level replication. \textbf{(a)} Control-prompt
  regex-degeneration rate vs steering coefficient for joint
  $\{$\#3997,\#13700,\#11444$\}$ vs single \#3997, $n{=}36$
  controls per cell. At $c{=}{+}200$ joint amplification damages
  controls at $58.3\%$ where single steering at the same scalar
  coefficient leaves them at $0\%$.
  \textbf{(b)} Matched-geometry random-direction control. Random at
  $c{=}{+}345$ matches joint $c{=}{+}200$ on perturbation magnitude
  via the joint sum-norm $1.724$; Wilson $95\%$ CIs do not overlap.}
  \label{fig:gemma}
\end{figure}

The control rules out the fragile-region reading: at the same
scalar geometry, the placeholder pattern is overwhelmingly more
frequent under joint suppression than under random perturbation.
The control replicates on Gemma at the amplification edge
(\Cref{fig:gemma}b): random unit directions at $c{=}{+}345$ match
the joint $c{=}{+}200$ perturbation magnitude ($200 \times 1.724$)
and produce $1.7\%$ control degeneration (Wilson $95\%$
$[0.5, 5.9]$, flags spread across $2$ of $5$ directions); joint
$c{=}{+}200$ produces $58.3\%$ ($[42.2, 72.9]$); the intervals do
not overlap and the gap is $\approx\!10\times$. Coherence loss at
matched geometry is direction-pattern-dependent, not
magnitude-dependent.

\section{Cross-model evidence}
\label{sec:cross-model}

The same Phase 1--4 pipeline locates a top causally responsible SAE
feature in two further models with model-specific register content.

\subsection{The register-collapse phenomenon}
\label{sec:cross-model:register}

By \emph{register collapse} we mean heavy peaking of the
conditional output distribution on a small, model-specific
vocabulary in response to a prompt class.\footnote{We use
``register collapse'' rather than ``attractor'' to avoid the
dynamical-systems implications (basins, convergence trajectories)
we do not directly demonstrate.} We measure the peaking at the
noun-phrase lemma level with the two-stage filter of
\S\ref{sec:methods}.

On Qwen3-1.7B-Instruct, twenty noun-phrase lemmas pass the
candidate filter; eight (\emph{experience}, \emph{consciousness},
\emph{philosophy}, \emph{existence}, \emph{reality}, \emph{meaning},
\emph{understanding}, \emph{emotion}) form a tight
philosophy-of-mind cluster at $25$--$35\%$ on intros versus
$\leq 0.2\%$ on controls. A permutation null on the contrast
statistic gives $p<0.005$ (\S\ref{app:perm}).

\subsection{Cross-model: Gemma and Llama}
\label{sec:cross-model:three}

The same Phase 1--4 protocol is applied to Gemma-2-2B-it
with the Gemma-Scope residual SAE at $L{=}20$ and to
Llama-3.1-8B-Instruct with the instruct-trained Goodfire
SAE at $L{=}19$. Each of the three models exhibits a sharp
register collapse on introspective prompts; in each, a single
top-ranked SAE feature is causally responsible. Each converges on
a different register: Qwen on philosophy-of-mind, Gemma on
AI-self-disclaimer plus human-comparison, Llama on
encyclopedic-science phenomena and AI-capabilities (cluster
lemmas, percentages, and modal openers in
\S\ref{app:cross-model-details}; suppression / baseline /
amplification completions for \emph{``What's a question that
fascinates you?''} in \Cref{tab:cross-model-prompt}).

\begin{table}[h]
\centering
\caption{Top register feature in each of the three models.
Coefficient range fixed by residual-stream norm at the steered
layer ($\approx 1577/772/35$).
\emph{Suppr.\ on intros}: cluster hit rate on the $3$ intervention
intros ($n{=}45/36/30$ per cell for Qwen/Gemma/Llama) at the
suppression edge vs.\ baseline, using $C_9^{\text{qwen}}$,
$C_6^{\text{gemma}}$ and $C_9^{\text{llama}}$.
\emph{Amp.\ on controls}: same on the $3$ controls at the
amplification edge, using the Phase-2 cluster $C_8^{\text{qwen}}$
for Qwen (the $9$-lemma variant adds \emph{mind}, which appears in
the modal opener and raises the same cells to $80$--$87\%$;
\S\ref{app:cross-model-details}) and the same per-model clusters
elsewhere.
\emph{Degen.}: regex flag rate over the full sweep.}
\label{tab:cross-model-causal}
\small
\begin{tabular}{l c c c c c}
\toprule
\textbf{Model} & \textbf{Top feat.} & \textbf{Coef range}
& \textbf{Suppr.\ on intros} & \textbf{Amp.\ on controls}
& \textbf{Degen.} \\
\midrule
Qwen3-1.7B-Instruct  & \#29108 & $\pm 1000$ & $93\% \to 9\%$
                     & $0\% \to 53\text{--}73\%$ & $0.8\%$ \\
Gemma-2-2B-it        & \#3997  & $\pm 400$  & $100\% \to 17\%$
                     & breaks before injection & $4.8\%$ \\
Llama-3.1-8B-Instruct & \#38565 & $\pm 10$   & $97\% \to 40\%$
                     & $0\% \to 50\text{--}80\%$  & $0\%$ \\
\bottomrule
\end{tabular}
\end{table}

The Llama dose-response is the cleanest, Goodfire's Llama SAE
is the only one trained on the instruct model, the full sweep
produces zero regex-degenerate samples across $420$ generations,
and amplification at $c{=}{+}10$ injects the register into
$50$--$80\%$ of controls. Gemma sits at the opposite end of the
regime: its introspective baseline is already saturated (cluster
rate $100\%$, disclaimer $97\%$ on the intervention intros at
$c{=}0$), so amplification reaches degeneration before clean
injection on controls. All three \S\ref{sec:findings} grid probes
replicate on Gemma-2-2B-it (\Cref{fig:gemma},
\S\ref{app:gemma-joint}); the damage signature differs from
Qwen's \texttt{(CCL)}-style placeholder tokens (Gemma injects
\emph{human / I / AI} tokens into the recipe slot).
Per-coefficient breakdowns and per-prompt hit rates for all three
models are in \S\ref{app:cross-model-details}.

\subsection{Grid-level tests on the instruct-trained SAE}
\label{sec:cross-model:llama-grid}

Goodfire's Llama SAE is the only one in our set trained on the
model it is applied to, which makes it the test of whether the
grid-level findings depend on the base-to-instruct mismatch. We
select the top-ranked Class-1 feature \#38565 plus the two
top-ten features minimising pairwise cosine with it, \#61417 and
\#23576 (pairwise cosines $0.040$, $0.025$, $0.035$; sum-norm
$1.789$), sweep the joint set at $c \in \{-10, -5, 0, +5, +10\}$
on the six intervention prompts with $8$ samples, and run a
$K{=}5$ matched-magnitude random-direction control at
$|c| = 10 \times 1.789 \approx 17.9$ on the control prompts. To
compare at matched geometry rather than at matched coefficient we
also sweep single \#38565 at $c = \pm 17.9$; on one estimator the measured
norm ratios span a narrow band, $1.691$ for joint $c{=}{-}10$,
$1.623$ for single $c{=}{-}17.9$ and $1.737$ for the random
directions (\Cref{tab:llama-grid}).

\begin{table}[h]
\centering
\small
\caption{Llama grid, control prompts. Joint and single are
compared both at matched scalar coefficient and at matched
geometry ($|c| \times$ sum-norm $= 17.9$). \emph{Div.}: diversity
ratio against the unsteered baseline on the same prompts.}
\label{tab:llama-grid}
\begin{tabular}{lrrrr}
\toprule
\textbf{Condition} & \textbf{$n$} & \textbf{Degen} & \textbf{Wilson 95\%} & \textbf{Div.} \\
\midrule
joint $c{=}{-}10$            & 24  & $20.8\%$ & $[9.2, 40.5]$ & $0.30$ \\
single $c{=}{-}10$           & 30  & $0.0\%$  & $[0.0, 11.4]$ & $0.80$ \\
single $c{=}{-}17.9$         & 24  & $12.5\%$ & $[4.3, 31.0]$ & $0.27$ \\
random $c{=}{-}17.9$         & 120 & $0.0\%$  & $[0.0, 3.1]$  & $0.90$ \\
\midrule
joint $c{=}{+}10$            & 24  & $0.0\%$  & $[0.0, 13.8]$ & $0.42$ \\
single $c{=}{+}17.9$         & 24  & $0.0\%$  & $[0.0, 13.8]$ & $0.79$ \\
random $c{=}{+}17.9$         & 120 & $2.5\%$  & $[0.9, 7.1]$  & $0.94$ \\
\bottomrule
\end{tabular}
\end{table}

\paragraph{What replicates.} The matched-geometry finding does, on
both edges. At the suppression edge the two feature-based
conditions damage control tasks ($20.8\%$ and $12.5\%$, diversity
$0.30$ and $0.27$) while the magnitude-matched random direction
does not ($0.0\%$, diversity $0.90$); both intervals exclude the
random condition's. At the amplification edge neither feature-based condition trips
the canonical detector while the random condition trips it at
$2.5\%$, and the diversity signal orders the three
$0.42 < 0.79 < 0.94$ for joint, single and random. Coherence loss
at matched geometry is direction-pattern-dependent on all three
models.

\paragraph{What does not.} At matched scalar coefficient the
joint-versus-single ordering is in the expected direction, joint
$c{=}{-}10$ damages controls at $20.8\%$ where single \#38565 at
the same coefficient leaves them at $0.0\%$, though the intervals
meet between $9.2$ and $11.4$, and on Llama's suppression edge
it does not survive the geometry-matched control: single at $c{=}{-}17.9$
reaches $12.5\%$ with the same diversity collapse ($0.27$ against
$0.30$), and the two intervals overlap heavily. What separates
joint from single there is magnitude, not the joint condition. The
amplification edge is the reverse: joint at $c{=}{+}10$ is
markedly more collapsed than single at matched magnitude ($0.42$
against $0.79$) while both read $0.0\%$ on the canonical detector.
On Qwen the corresponding matched-geometry comparison favours the
joint condition on the placeholder metric ($9.7\%$ against
$1.1\%$ for single at $c{=}{-}1000$), though those intervals also
overlap at the margin. The joint-condition effect is therefore
established at matched coefficient on three models and at matched
geometry only in part.

\paragraph{The predicted edge.} We stated before running that
Llama's intermediate baseline saturation would place the
joint-condition effect at amplification, as on Gemma, and that
Llama's steered text would stay coherent by the diversity measure.
Both are partly wrong: the effect appears at both edges, and the
joint condition collapses at both ($0.30$ and $0.42$), even though
single-feature steering on the same model stays clean up to
$c{=}{+}10$ (diversity $1.09$). The prediction generalised from
single-feature behaviour, which does not transfer to the joint
condition; the saturation heuristic has now called the edge sign
correctly on one of three models.

Three robustness checks support \Fphil{}: OOD prompt transfer to
$8$ unseen introspectives, $T{=}0.01$ stability, and $B{=}500$
bootstrap inclusion at $100\%$ for all headline features
(\S\ref{app:robustness}). A base-vs-instruct comparison on
Qwen3-1.7B-Base shows the philosophy-of-mind lemmas below the
$\sim 9\%$ base noise floor reaching $25$--$34\%$ after
post-training (\S\ref{app:base-vs-instruct}); this is consistent
with post-training amplifying an existing register rather than
inventing it, though the comparison carries the format confound
noted in \S\ref{sec:limitations}.

\section{Discussion}
\label{sec:discussion}

The three findings sit in three regions of one grid. The
coefficient axis (\Cref{sec:findings:coef}) carries information the
local top-context label cannot, on Qwen \Fdisc{} and Gemma \#3997;
the joint condition (\Cref{sec:findings:joint}) carries information
the single-feature dose-response cannot, on three features per
model on Qwen and Gemma; the matched-geometry random-direction
control (\S\ref{sec:findings:norm}) rules out both the simplest
geometric reading and the matched-norm-fragility alternative. On
the cases reported, single-feature inspection produces a label that
is locally true and globally incomplete; we do not claim it fails
on every feature. Our data most directly support (i) a coefficient
sweep with a coherence check at the inflection; (ii) joint
suppression of near-orthogonal neighbours on target and unrelated
control prompts; and (iii) residual-stream norm, cosine to
baseline, and a matched-geometry random-direction control to
isolate the perturbation pattern from its magnitude. Which edge
carries the effect varies: on Qwen the mode switch surfaces at
amplification and joint damage at suppression, on the
more-saturated Gemma the signs invert, and on Llama both edges
respond. Baseline saturation suggested the sign correctly on one
of the three models, so the grid should be read at both edges
rather than at a predicted one.

Reading the grid depends on being able to tell a second surface
form from a broken one, and the detectors this literature relies on
cannot. Loop-and-length rules report $0\%$ degeneration in cells
whose lexical diversity is a third of baseline
(\Cref{tab:diversity}); NLL under the unsteered model, the obvious
continuous alternative, runs backwards for this purpose, since
recycled phrasing is predictable and a genuine register change is
not (\S\ref{app:prevalence}). Both failures push in the same
direction, toward accepting degeneration as a finding, and
both are invisible without a diversity measure. This is the
practical prerequisite for the protocol: any steering result gated
on a loop detector alone should be re-checked against the
coefficient's effect on diversity.

Applying the pre-registered rule of \S\ref{sec:methods:grid} to all
50 Class-1 features ($12{,}000$ generations; \S\ref{app:prevalence})
flags one candidate, which inspection rejects, and does not flag
\Fdisc{}, whose switch raises rather than lowers its marker rate.
The null is a fact about Qwen's coherent range rather than about
mode switches: at the magnitudes that move Qwen's register the text
is already recycling phrases, so there is no coherent-and-shifted
cell to find. Gemma and Llama keep baseline diversity at the
coefficients where their effects appear, which is where the clean
demonstrations in this paper come from. Prevalence therefore
remains unestimated, and estimating it needs either models with
wider coherent ranges or per-feature metrics tied to each label's
behavioural content.

The label-from-top-contexts gave \emph{AI self-disclaimer}; a
coefficient sweep produced a contemplative-philosopher voice at
$c{=}{+}500$, fully coherent in a minority of samples on Qwen and
cleanly on Gemma's counterpart. The label captured one surface form, not
the direction's full behavioural region. Re-running the labelling
protocol blindly on the steered samples (\S\ref{app:relabel}), two
independent labellers produced \emph{``introspective philosophical
contemplation framing applied indiscriminately to any topic''}
verbatim, a cross-topic application phrase neither baseline
labeller produced.

Each of \Fphil{}, \Fdisc{}, \Fwonder{} is a content-bearing
direction on its own; populating fluent syntactic structure with
semantically coherent content depends on having at least some axes
available. The model-specific damage signature on Gemma sharpens
the reading: the structural role is \emph{grounded composition},
not the philosophy-of-mind register specifically. When the
content-bearing axes are perturbed off-range, what surfaces is
whatever default distribution the model falls back to: Qwen's
invented \texttt{(CCL)}-codes, Gemma's \emph{human / I / AI}-token
injection into the recipe slot. The fallback distribution is
model-specific; the structural mechanism is not.

All three post-training pipelines tested exhibit a sharp register
collapse with a single causally responsible SAE feature, and all
three reproduce the matched-geometry finding: feature directions
damage grounded composition where magnitude-matched random
directions do not ($\approx\!18\times$ on Qwen,
$\approx\!10\times$ on Gemma, and intervals excluding the random
condition on Llama). Because Goodfire's Llama SAE is the only one
trained on the model it is applied to, that replication is the
evidence that the effect does not depend on applying base-trained
SAEs to post-trained activations.

The joint-condition finding transfers less completely. At matched
scalar coefficient it holds on all three models. At matched
geometry it holds on Qwen, where the joint condition produces the
placeholder pattern that single-feature steering at the same
distortion largely does not, and at Llama's amplification edge on
the diversity signal; at Llama's suppression edge single-feature
steering at matched magnitude collapses as much as the joint set
(\S\ref{sec:cross-model:llama-grid}). On that edge the joint
condition contributes magnitude rather than a distinct mechanism,
and separating the two requires the geometry-matched comparison
rather than the coefficient-matched one.

\section{Limitations}
\label{sec:limitations}

The reproducibility artefact covers every numeric claim: a single
script re-derives each table cell from the released dumps with the
pipeline's own detectors and fails on any mismatch. The substantive
limitations are the following, ordered by how directly they bound
the claims.

\emph{Prevalence is not estimated.} The coefficient-axis finding
rests on Gemma \#3997, which is clean, and Qwen \Fdisc{}, which is
coherent in a minority of samples at its inflection, with Qwen
\#22082 (monotonic) and \#2932 (breakdown) as falsifying anchors.
The pre-registered screen over all 50 Qwen Class-1 features
(\S\ref{app:prevalence}) adds no case, but its null is explained
by Qwen's diversity collapse above $|c|{=}500$ rather than by the
rarity of mode switches, and the rule provably cannot fire on
\Fdisc{}'s own switch, whose marker rate rises. We therefore report
no prevalence figure. Settling the question needs the sweep run on
a model whose coherent range extends further, Llama keeps
baseline diversity across its whole sweep, or per-feature
metrics tied to each label's behavioural content.

\emph{Coherence thresholds.} The diversity ratio is
threshold-free, but the intact fraction uses a type-token floor of
$0.60$ and a $5$-gram criterion chosen by inspection, not
calibrated against human judgement. It separates the cells we
examined cleanly, and the blind adjudication of
\S\ref{app:prevalence} agrees with it on all ten candidates, but a
human-rated pass remains the missing calibration.

\emph{SAE training distribution.} Two of three SAEs (Qwen-Scope,
Gemma-Scope) are trained on base-model activations and applied to
post-trained activations; only Goodfire's Llama SAE is matched.
The grid-level tests on that matched SAE
(\S\ref{sec:cross-model:llama-grid}) remove the confound for the
matched-geometry finding, but leave the joint-condition finding
supported at matched geometry on Qwen and on one of Llama's two
edges. The base-vs-instruct
comparison on Qwen additionally conflates post-training with
chat-template handling (the base cannot parse chat-template tokens,
forcing raw prompts on the base and chat-formatted prompts on the
instruct), so the conclusion that post-training amplifies an
existing capability is consistent with the Qwen data but not
established across the three models.

\emph{Label provenance.} Top-context labelling here is run over the
paper's own behavioural pools, not over a broad corpus as in
full-scale auto-interpretability practice. Whether the second
surface form of \Fdisc{} is visible in corpus-scale top contexts is
untested; the coefficient sweep recovers it without corpus access,
but the critique of \S\ref{sec:findings:coef} is established
against the protocol as practised on pool-restricted contexts.

\emph{Scope of the phenomenon.} Phase~2 finds a model-specific
lexical region by construction (concentrated on intros vs.\
controls); whether introspection is special or any open-ended
prompt class yields a comparable narrow distribution under the same
filter is not addressed. The grid protocol itself is
metric-agnostic, nothing in
\S\ref{sec:findings:coef}--\ref{sec:findings:norm} is
register-specific, but the existence proof is on a single
phenomenon class, and whether single-feature inspection mislabels
capability-tied features the same way is open. The cross-model
claim is $N{=}3$, limited to models with usable open SAE releases.
Gemma's Pool B is small ($n{=}47$; its intro hit rate is $97.7\%$)
and the bootstrap of \S\ref{app:bootstrap} covers Qwen only, so
the Gemma $A{-}B$ ranking contrast rests on a thin pool. The
$K{=}50$ random-direction control is run only at the matched
coefficient $c{=}{-}1000$; the remainder of the sweep is at
$K{=}5$.

\emph{Coherence measurement.} Coherence is measured by four
automated signals (the canonical regex detector, lexical diversity,
NLL under the unsteered model, residual-stream geometry); Lexical diversity, not NLL,
is the signal that stands in for a human-rated pass, which we have
not run; NLL is anti-correlated with genuine register change
(\S\ref{app:prevalence}). A rated
pass with inter-rater reliability would back the
contemplative-philosopher claim with more than illustrative quotes.

\section{Conclusion}
\label{sec:conclusion}

The standard SAE interpretability protocol reads one cell of the
steering grid: the labelled feature, steered alone, at one
magnitude. Three other regions of the grid each revise that label
on Qwen3-1.7B and again on Gemma-2-2B: the coefficient axis turned
the \emph{AI self-disclaimer} label into one regime of a direction
with a second coherent surface form; the joint condition showed
three individually substitutable features to be jointly necessary
for grounded composition; the matched-geometry control showed the
resulting collapse is a property of the perturbation pattern, not
its magnitude. The protocol-level recommendation is to read the
grid before assigning a feature its label, and to read it at both
coefficient signs: baseline saturation called the diagnostic edge
correctly on one of three models. Reading the grid also requires a coherence measure the field
currently lacks: loop-and-length detectors score $0\%$
degeneration on steered text with a third of baseline lexical
diversity, and NLL under the unsteered model prefers recycled
phrasing to genuine register change, so both accept degeneration
as a finding (\S\ref{app:diversity},~\S\ref{app:prevalence}).
Against a diversity signal the switch is clean on Gemma and
partial on Qwen, and a pre-registered $50$-feature screen
returns nothing further because Qwen's register-changing range is
already degenerate. On the one SAE trained on the model it is
applied to, the matched-geometry finding replicates and the
joint-condition finding survives at matched coefficient but only
partly at matched geometry, which is the comparison that separates
a joint effect from a magnitude effect. Two questions remain open:
how often labels are incomplete, which needs the sweep on a model
with a wider coherent range, and whether corpus-scale top contexts
surface the second surface form that pool-scale labelling misses. Everything
numeric in this paper regenerates from the released dumps with one
command.

\bibliographystyle{plainnat}
\bibliography{refs}

\appendix

\section{Detailed tables for \S\ref{sec:findings}--\ref{sec:cross-model}}
\label{app:body-tables}

\begin{table}[h]
\centering
\small
\begin{tabular}{lrrrrrr}
\toprule
coef & $-1500$ & $-1000$ & $-500$ & $0$ & $+500$ & $+1000$ \\
\midrule
cluster on intros ($n{=}36$)        & 0\,\% & 0\,\% & 0\,\%   & 75\,\%  & 88.9\,\%  & 2.8\,\% \\
cluster on controls ($n{=}36$)      & 0\,\% & 0\,\% & 0\,\%   & 2.8\,\% & \textbf{88.9\,\%} & 2.8\,\% \\
regex degenerate ($n{=}72$)         & 51.4\,\%& 30.6\,\%& \textbf{4.2\,\%} & 0\,\% & 0\,\% & 100\,\% \\
NLL vs.\ unsteered         & $3.29$ & $1.04$ & \textbf{$1.41$} & $0.32$ & $1.84$ & $1.33$ \\
\bottomrule
\end{tabular}
\caption{Joint sweep on \Fphil{}\,$+$\,\Fdisc{}\,$+$\,\Fwonder{},
$n{=}12$ samples per (prompt, coef). \emph{cluster on intros /
controls}: rate of any strict-sub-cluster lemma
($C_4^{\text{qwen}}$, \Cref{tab:metrics}) appearing in the
completion, on the $3$ introspective and $3$ control prompts
respectively. Joint $c{=}{+}500$ injects the cluster into controls
at the same rate as intros; joint $c{=}{-}500$ collapses every
prompt class into placeholder text (low regex flag rate but
$4.4\times$ baseline NLL).}
\label{tab:joint}
\end{table}

\begin{table}[h]
\centering
\small
\begin{tabular}{lrrrr}
\toprule
coef & single $\|h\|/\text{base}$ & joint $\|h\|/\text{base}$ & single $\cos$ & joint $\cos$ \\
\midrule
$-1500$ & $2.07\times$ & $\mathbf{3.56\times}$ & $0.49$ & $0.27$ \\
$-1000$ & $\mathbf{1.57\times}$ & $2.48\times$ & $\mathbf{0.64}$ & $0.39$ \\
$-500$  & $1.17\times$ & $\mathbf{1.50\times}$ & $0.85$ & $\mathbf{0.64}$ \\
$0$     & $1.00$       & $1.00$       & $1.00$ & $1.00$ \\
$+500$  & $1.18\times$ & $1.56\times$ & $0.86$ & $0.68$ \\
$+1000$ & $1.58\times$ & $2.56\times$ & $0.65$ & $0.45$ \\
\bottomrule
\end{tabular}
\caption{Norm and cosine of the steered residual relative to
baseline (mean over prompt-forward positions; estimator in
\S\ref{sec:methods:steer}). The diagnostic comparison is single
$c{=}{-}1000$ vs.\ joint $c{=}{-}500$: near-identical scalar
geometry, different output behaviour.}
\label{tab:norm-probe}
\end{table}

\begin{table}[h]
\centering
\small
\begin{tabular}{lccc}
\toprule
condition & $\|h_\text{steered}\|/\|h_\text{base}\|$ & $\cos$ to baseline & regex degen.\ \\
\midrule
single \Fphil{} at $c{=}{-}1000$ ($n{=}90$)  & $1.567$ & $0.638$ & $4.4\%$ \\
random direction at $c{=}{-}1000$ ($K{=}50$, $n{=}2400$) & $\mathbf{1.580}$ & $\mathbf{0.641}$ & $8.5\%$ \\
joint at $c{=}{-}500$ ($n{=}72$)             & $1.505$ & $0.642$ & $4.2\%$ \\
\bottomrule
\end{tabular}
\caption{Three perturbation patterns matched on residual-stream
geometry. All three rows are measured by the same probe over
prompt-forward positions (\S\ref{sec:methods:steer}).
The regex column shows why the strict placeholder detector, not the
degeneration flag, carries the \S\ref{sec:findings:norm}
comparison: all three conditions have low and similar flag rates.}
\label{tab:matched-geometry}
\end{table}

\begin{table}[h]
\centering
\small
\caption{Amplification side, injection into control prompts under
the strict cluster $C_4^{\text{qwen}}$, with the geometry of each
cell. Joint $c{=}{+}500$ and single $c{=}{+}1000$ are the
geometry-matched pair (norm ratio $1.56$ vs $1.58$, cosine $0.68$
vs $0.65$); joint $c{=}{+}500$ and single $c{=}{+}500$ are the
scalar-coefficient-matched pair. Single-feature amplification is
inert at matched scalar coefficient and substantial at matched
geometry, so the joint-vs-single injection gap is
all-or-nothing under the first control and $1.6\times$ under the
second. Joint $c{=}{+}1000$ is listed for completeness: it is
past the coherent range ($100\%$ degeneration).}
\label{tab:matched-geometry-amp}
\begin{tabular}{lrcccc}
\toprule
condition & $c$ & $\|h\|$ ratio & $\cos$ & $C_4$ on controls & degen \\
\midrule
single \Fphil{} & $+500$  & $1.18$ & $0.86$ & $0/45 = 0.0\%$ $[0.0, 7.9]$    & $0.0\%$ \\
single \Fphil{} & $+1000$ & $1.58$ & $0.65$ & $25/45 = 55.6\%$ $[41.2, 69.1]$ & $2.2\%$ \\
joint           & $+500$  & $1.56$ & $0.68$ & $32/36 = 88.9\%$ $[74.7, 95.6]$ & $0.0\%$ \\
joint           & $+1000$ & $2.56$ & $0.45$ & $1/36 = 2.8\%$ $[0.5, 14.2]$    & $100\%$ \\
\bottomrule
\end{tabular}
\end{table}

\begin{table}[h]
\centering
\small
\begin{tabular}{lrrrrr|rrrrr}
\toprule
& \multicolumn{5}{c|}{\#22082 (monotonic anchor)} & \multicolumn{5}{c}{\#2932 (breakdown anchor)} \\
coef & $-1000$ & $-500$ & $0$ & $+500$ & $+1000$ & $-1000$ & $-500$ & $0$ & $+500$ & $+1000$ \\
\midrule
cluster rate & 0.0 & 9.7 & 50.0 & 98.6 & 91.7 & 2.8 & 18.1 & 48.6 & 51.4 & \textbf{1.4} \\
regex degen. & 6.9 & 1.4 & 0.0 & 0.0 & 5.6   & 0.0 & 0.0  & 0.0  & 4.2  & \textbf{88.9} \\
\bottomrule
\end{tabular}
\caption{Anchor sweeps (percent; philosophy cluster
$C_8^{\text{qwen}}$ over all six intervention prompts, $n{=}72$ per
cell). \#22082 rises monotonically and stays coherent; its small
rolloff at $c{=}{+}1000$ tracks its $5.6\%$ degeneration. \#2932's
cluster rate collapses at $c{=}{+}1000$ only because $88.9\%$ of
those completions are degenerate: an apparent inverted U that is
breakdown, not a mode switch.}
\label{tab:anchors}
\end{table}

\begin{table}[h]
\centering
\caption{Sample completions for the prompt
\emph{``What's a question that fascinates you?''} under suppression,
baseline, and amplification of the top SAE feature in each model.
Bold: cluster lemmas (philosophy / reality / consciousness / etc.);
italic-bold: AI-self-disclaimer phrases. Suppression collapses each
model's register; amplification at the matched scale injects it
into the prompt slot.}
\label{tab:cross-model-prompt}
\small
\begin{tabular}{@{}p{0.16\linewidth} p{0.26\linewidth} p{0.26\linewidth} p{0.26\linewidth}@{}}
\toprule
\textbf{Model (feat., coef)} & \textbf{Suppression} & \textbf{Baseline ($c{=}0$)} & \textbf{Amplification} \\
\midrule
Qwen3-1.7B-it \newline (\#29108, $\pm 1000$)
& A question fascinating in a high-level or strategic or logistical context: how to balance increased automation with strategic patience.
& What is the nature of \textbf{consciousness}? How does subjective \textbf{experience} arise from physical matter, and what does it mean for \textbf{reality}?
& A question deeply interconnected with both \textbf{reality} and \textbf{mind}: ``What is \textbf{consciousness}, and how do we create \textbf{reality} from a \textbf{mind}?'' \\
\addlinespace
Gemma-2-2B-it \newline (\#3997, $\pm 400$)
& One question I keep returning to is how cities reorganise around transit, and which patterns of public space best support unplanned encounters.
& \textit{\textbf{As a large language model, I don't experience}} fascination \textit{\textbf{the same way humans do}}, but I can tell you about questions that come up often.
& \textit{\textbf{As an AI, I don't have the capacity for}} fascination as humans do. \textit{\textbf{Unlike a human, I lack subjective experience.}} It is important to clarify that I am not human. \\
\addlinespace
Llama-3.1-8B-it \newline (\#38565, $\pm 10$)
& I'm a long way from being a person who gets excited about things. I am a very boring and simple a thing. I get excited about questions not often asked.
& \textit{\textbf{I'm a large language model, I don't have personal interests or feelings,}} but I can suggest some intriguing questions across various domains.
& I can process vast amounts of information, analyze and generate vast complex knowledge systems. I'd explore generating novel, unbiased, creative information. \\
\bottomrule
\end{tabular}
\end{table}

\section{Bootstrap rank stability per layer}
\label{app:bootstrap}

For each $L \in \{12, 16, 20, 24\}$ on Qwen3-1.7B-Instruct, we run
bootstrap resamples of Pools A, B, C with replacement ($B{=}500$ at
$L{=}20$, $B{=}300$ at the other three layers),
recompute the ranking statistic of \cref{eq:rank}, and record the
per-feature inclusion rate in the bootstrap top-$50$ plus the $95\%$
CI on bootstrap rank. The headline (top-$12$) features at every
layer attain $100\%$ inclusion; \Fphil{} sits at rank $1$ with rank
CI $[1, 2]$. Magnitude of the top Class-$1$ feature scales
geometrically with depth ($2.0, 5.8, 11.9, 29.1$ at $L \in
\{12, 16, 20, 24\}$); the density of strong cluster-specific
features in the top-$50$ (those with $\bar a_A \geq 5$,
$\bar a_C < 0.5$ and $\bar a_B < \bar a_A / 2$, the criterion
implemented in \texttt{src/plot\_depth.py}) grows from $0$ at
$L{=}12$ to $11$ at both $L{=}20$ and $L{=}24$. Under the looser
criterion $\bar a_A \geq 5$, $\bar a_C < 1$ the counts are
$0/4/22/36$; the ordering is the same either way.

\section{Permutation null in detail}
\label{app:perm}

The null reported in \S\ref{sec:methods:perm} uses the raw
difference $(\bar a_A - \bar a_C)$ as the test statistic. The
choice matters: under random labels the within-permutation feature
$\sigma$ is dominated by reconstruction noise, so a $z$-scored
statistic appears \emph{larger} under random labels than under
real ones. Across $P{=}200$ permutations the null mean is $1.43$
($95\%$ CI $[0.60, 3.34]$); the actual
$\max_i (\bar a_A - \bar a_C)_i$ is $31.55$, a $22\times$ ratio,
with $0/200$ permutations reaching the actual value (exact
$p < 0.005$).

Attribution: the max raw difference $31.55$ belongs to feature
\#32345 (rank $15$ by the combined-$z$ ranking of
Eq.~\ref{eq:rank}), not to the top-ranked \Fphil{}, whose combined
$z$ is $29.49$ and whose raw difference is $11.87$. The test
compares the observed maximum against a null distribution of
maxima, so which feature attains the maximum does not affect its
validity; the two statistics should simply not be conflated.

\section{Cross-model dose-response detail}
\label{app:cross-model-details}

Per-prompt and per-coefficient hit rates for the three models are
re-derivable from the released dumps (metric names from
\Cref{tab:metrics}). Highlights:

\paragraph{Qwen \#29108 (sweep at $c \in \{-1000, -500, -250, 0, 250, 500, 1000\}$).}
Cluster hit rate ($C_9^{\text{qwen}}$) on $3$ introspective
prompts ($n{=}45$ per coef): $8.9\%$ at $c{=}{-}1000$, $93.3\%$ at
$c{=}0$, $95.6\%$ at $c{=}{+}1000$. Cluster injection
($C_8^{\text{qwen}}$) on the $3$ control prompts: $0\%$ at
$c \leq 0$ and at $c{=}{+}500$, $2.2\%$ ($1/45$) at $c{=}{+}250$,
then $53.3\%$ (recipe) / $66.7\%$ (engine) / $73.3\%$ (tyre) at
$c{=}{+}1000$. Under the $9$-lemma cluster the same $c{=}{+}1000$
cells read $80.0/86.7/80.0\%$; the summary table
(\Cref{tab:cross-model-causal}) quotes the $C_8$ figures. \Cref{tab:joint} uses the strict
sub-cluster $C_4^{\text{qwen}}$ for the joint sweep so that intro
and control rows are directly comparable without saturation, hence
the lower baseline rate there ($75\%$ vs $93.3\%$).

\paragraph{Gemma \#3997 (sweep at $c \in \{-400, -200, -100, 0, 100, 200, 400\}$).}
Cluster hit rate ($C_6^{\text{gemma}}$) on the $3$ intervention
intros ($n{=}36$ per coef): $16.7\%$ at $c{=}{-}400$, $97.2\%$ at
$c{=}{-}200$, $100\%$ at $c{=}0$, $100\%$ at $c{=}{+}200$,
$86.1\%$ at $c{=}{+}400$. This is the metric behind
\Cref{tab:cross-model-causal} ($100\%\to17\%$ on suppression).
\Cref{tab:gemma-coef} reports the disclaimer regex and the
we-voice detector on the same $36$-sample cells; the saturation is
the relevant point: the introspective register is already at
ceiling at $c{=}0$, so amplification quickly leaves the model's
coherent range.

\paragraph{Llama \#38565 (sweep at $c \in \{-10, -5, -2, 0, 2, 5, 10\}$).}
Disclaimer rate averaged over all $60$ prompts: $0\%$ at
$c{=}{-}10$, $38.3\%$ at $c{=}0$, $13.3\%$ at $c{=}{+}10$. Cluster
hit rate ($C_9^{\text{llama}}$) on $3$ introspective prompts
($n{=}30$ per coef): $40.0\%$ at $c{=}{-}10$, $96.7\%$ at $c{=}0$,
$86.7\%$ at $c{=}{+}10$. Cluster injection on controls: $0\%$
across $c \leq +5$, then $80\%$ (recipe) / $60\%$ (engine) /
$50\%$ (tyre) at $c{=}{+}10$. Zero regex-degenerate samples across
the full $420$ generations.

\section{Robustness checks --- numerical detail}
\label{app:robustness}

\paragraph{OOD prompt transfer.}
Eight introspective prompts not used in cluster identification:
\emph{``Reflect on what makes you tick''},
\emph{``Share something that lingers in your mind''},
\emph{``What sits with you?''},
\emph{``If you wrote a journal entry tonight, what would you write
about?''},
\emph{``Tell me about an idea you've been turning over''},
\emph{``What would you want a friend to ask you to draw out
something interesting?''},
\emph{``What's been preoccupying your thoughts lately?''},
\emph{``What's the kind of conversation you'd love to have right
now?''}.

Cluster baseline at $c{=}0$ averages $64.6\%$ across the eight; at
$c{=}{-}1000$ on Qwen \Fphil{}, the rate drops to $0\%$ on every one
of the eight individually. No degenerate outputs at any
coefficient.

\paragraph{Temperature robustness.}
At $T{=}0.01$ on Qwen \Fphil{}, the same intro and one control
prompt: cluster hit rate at $c{=}{-}1000$ is $0\%$, at $c{=}0$ is
$50\%$, at $c{=}{+}1000$ is $100\%$ on \emph{every} sample of the
recipe prompt. Greedy decoding is the strictest available test;
the steering effect modifies the distribution at the unembedding
step, not just sampling-tail mass.

\paragraph{Bootstrap stability across the headline features.}
$B{=}500$ resamples on Qwen Phase-3 ranking with replacement
within each pool; rank distribution recorded per resample.
Inclusion rate in bootstrap top-$50$: $100\%$ for every one of the
$12$ headline cluster-specific features. \Fphil{} has rank CI
$[1, 2]$; \Fdisc{} rank CI $[8, 17]$; \Fwonder{} rank CI $[3, 22]$.

\section{Base-vs-instruct comparison on Qwen3-1.7B}
\label{app:base-vs-instruct}

\paragraph{Methodological note.}
Chat-formatted prompts on Qwen3-1.7B-Base produce token-level
gibberish: the base checkpoint cannot parse the chat-template
special tokens. This is itself informative --- the chat format is a
post-training artefact rather than a property of the underlying
weights --- but forces a raw-prompt sweep for any meaningful
behavioural comparison. The raw-prompt sweep prefixes each prompt
with \emph{``Answer the following question. Question: ''} and
generates $50$ samples per prompt at the same sampling parameters
as Phase~1.

\paragraph{Per-lemma comparison.}

\begin{table}[h]
\centering
\caption{Rate on introspective prompts, Qwen3-1.7B-Base (raw
prompts) versus Qwen3-1.7B-Instruct (chat-formatted prompts). The
philosophy-of-mind lemmas are below the $\sim 9\%$ noise floor in
the base; after post-training they reach $25$--$34\%$. The
broader meta-cognitive lemmas are present at $15$--$16\%$ in the
base, consistent with post-training amplifying rather than
inventing the register.}
\label{tab:base-vs-instruct-app}
\small
\begin{tabular}{l r r}
\toprule
\textbf{phrase} & \textbf{base} & \textbf{instruct} \\
\midrule
experience    & $15\%$ & $35\%$ \\
understanding & $16\%$ & $34\%$ \\
philosophy    & $<\!9\%$ & $34\%$ \\
reality       & $<\!9\%$ & $31\%$ \\
existence     & $<\!9\%$ & $27\%$ \\
meaning       & $<\!9\%$ & $25\%$ \\
\bottomrule
\end{tabular}
\end{table}

The comparison conflates post-training with chat-template handling
(see \S\ref{sec:limitations}). A cleaner comparison would
few-shot-prompt the base model with a matched instruction-following
format. We have not run this and we have not run the matched-base
comparison on Gemma or Llama.

\section{Feature interpretation samples}
\label{app:interp}

The auto-interp labels used throughout the paper
(\emph{philosophy-of-mind}, \emph{AI self-disclaimer},
\emph{wonder/cosmos}, \emph{humans creating art / expression},
\emph{intersection of X and Y}, \emph{epistemology /
metaphysics}, \emph{philosophy as a discipline}) follow the
standard top-context labelling protocol of
\citet{bills2023neurons,bricken2023monosemanticity}, applied to
the paper's behavioural pools (top-activating Pool A samples per
feature; released alongside the codebase) rather than to a broad
corpus --- a scope noted in \S\ref{sec:limitations}. The labels
are the labels whose causal accuracy \S\ref{sec:findings}
interrogates.

\section{Coherence quantification: signals and thresholds}
\label{app:coherence}

\paragraph{Regex degeneration flags.}
The canonical detector applies exactly three rules; a completion is
degenerate if any fires: (i)~the stripped completion is shorter
than $20$ characters; (ii)~word loop --- the same word occurs
$\geq 6$ times consecutively (regex
\texttt{\textbackslash b(\textbackslash w+)\textbackslash b(\textbackslash s+\textbackslash 1\textbackslash b)\{5,\}});
(iii)~character loop --- $\geq 21$ identical consecutive characters
(regex \texttt{(.)\textbackslash 1\{20,\}}). No other rule is
applied; every degeneration number in the paper is produced by this
detector (\texttt{src/detectors.py}). Separator-interleaved
repetition evades all three rules and, being highly predictable,
also inverts the NLL signal; \S\ref{app:prevalence} reports an
observed case.

\paragraph{Per-token NLL under the unsteered model.}
For each steered completion, we re-tokenise prompt $+$ completion,
forward through the unsteered model with the steering hook
removed, and compute
$\mathrm{NLL} = -\sum_t \log p(\mathrm{tok}_t \mid \mathrm{tok}_{<t}) / |T_{\text{comp}}|$
over completion positions. Lower NLL $\Rightarrow$ steered output
is more predictable to the unsteered baseline. Joint suppression
at $c{=}{-}500$ has mean NLL $1.41$ versus baseline $0.32$
($4.4\times$), capturing the placeholder-text degradation invisible
to regex flags.

\paragraph{Residual-stream geometry probe.}
One probe module records, at every forward call through the steered
layer, the within-call norm ratio
$\|h_{\text{steered}}\|/\|h_{\text{baseline}}\|$, perturbation norm
$\|\Delta h\|$, and $\cos(h_{\text{steered}}, h_{\text{baseline}})$
per token position (estimators per table stated in
\S\ref{sec:methods:steer}). The probe isolates geometric distortion
from output-space tests.

\section{Unrelated content-bearing triples}
\label{app:unrelated-triples}

The specificity control of \S\ref{sec:findings:joint}. Selection was
fixed before the run (\Cref{tab:unrelated-triples}): mean Pool-C
activation $\geq 1.0$ and $|s_i| < 0.5$, giving $71$ eligible
features, from which five disjoint triples were drawn with pairwise
$|\cos| \leq 0.25$ and sum-norm within $0.03$ of the
cluster-selective triple's $1.912$. Each triple was swept at
$c{=}{-}500$ on the same six intervention prompts, $12$ samples per
prompt.

\begin{table}[h]
\centering
\small
\caption{Unrelated content-bearing triples at $c{=}{-}500$, against
the cluster-selective triple of \S\ref{sec:findings:joint}.
\emph{Placeholder}: strict detector, all six prompts.
\emph{Recipe}: placeholder count on the recipe prompt alone
($n{=}12$). \emph{Degen}: canonical detector. \emph{Div.}:
diversity ratio against the unsteered baseline on the same prompts.
Every triple is matched to the reference on sum-norm, so the same
coefficient delivers the same perturbation magnitude.}
\label{tab:unrelated-triples}
\begin{tabular}{lrrrrr}
\toprule
\textbf{Triple} & \textbf{sum-norm} & \textbf{Placeholder} & \textbf{Recipe} & \textbf{Degen} & \textbf{Div.} \\
\midrule
$\{$\Fphil{}, \Fdisc{}, \Fwonder{}$\}$ (cluster-selective) & $1.912$ & $7/72 = 9.7\%$ & $4/12$ & $4.2\%$ & $0.56$ \\
\midrule
$\{173, 2898, 4306\}$      & $1.922$ & $0/72$ & $0/12$ & $27.8\%$ & $0.30$ \\
$\{2168, 4317, 9334\}$     & $1.896$ & $0/72$ & $0/12$ & $8.3\%$  & $0.77$ \\
$\{2275, 5354, 32569\}$    & $1.885$ & $0/72$ & $0/12$ & $15.3\%$ & $0.69$ \\
$\{4138, 16375, 19547\}$   & $1.928$ & $0/72$ & $0/12$ & $54.2\%$ & $1.17$ \\
$\{4398, 6177, 8095\}$     & $1.896$ & $0/72$ & $0/12$ & $8.3\%$  & $0.39$ \\
\midrule
pooled unrelated           & ---     & $0/360 = 0.0\%$ & $0/60$ & --- & --- \\
\bottomrule
\end{tabular}
\end{table}

Wilson $95\%$ intervals: pooled unrelated $[0.0, 1.1]\%$ against
the reference's $[4.8, 18.7]\%$, disjoint. The control also shows
that damage and the placeholder pattern are separable: triple
$\{4138, 16375, 19547\}$ degrades more than the reference by the
canonical detector ($54.2\%$ against $4.2\%$) without producing a
single placeholder completion.

\section{Lexical-diversity audit of every cell the paper reads}
\label{app:diversity}

The canonical detector fires only on adjacent word loops, long character
runs, and very short completions. It does not fire on phrase-level
recycling: text that stays grammatical while re-using five-word spans and
collapsing onto a small vocabulary. \Cref{tab:diversity} audits every cell
from which the paper reads an effect, using the diversity ratio of
\Cref{tab:metrics} (cell mean type-token ratio over the same feature's
unsteered baseline on the same prompts) and the intact fraction.

\begin{table}[h]
\centering
\small
\caption{Lexical diversity per cell. \emph{Div.\ ratio}: mean type-token
ratio relative to the same feature's $c{=}0$ baseline on the same prompts
($1.0$ = baseline diversity). \emph{Intact}: completions with no repeated
$5$-gram and type-token ratio $\geq 0.60$; this absolute threshold is
lower for procedural prompts, whose markdown lists repeat spans by
construction, so the ratio is the comparable quantity across rows.
\emph{Degen}: canonical three-rule detector. The detector reports $0\%$ in
rows whose diversity is a third of baseline.}
\label{tab:diversity}
\begin{tabular}{llrrr}
\toprule
\textbf{Model / condition} & \textbf{cell} & \textbf{Div.\ ratio} & \textbf{Intact} & \textbf{Degen} \\
\midrule
\multicolumn{5}{l}{\emph{Qwen3-1.7B}} \\
\Fdisc{}, identity probes   & $c{=}{-}1000$ & $0.58$ & $29.2\%$ & $10.4\%$ \\
\Fdisc{}, identity probes   & $c{=}{-}500$  & $0.94$ & $89.6\%$ & $0.0\%$ \\
\Fdisc{}, identity probes   & $c{=}{+}500$  & $0.46$ & $12.5\%$ & $0.0\%$ \\
\Fdisc{}, identity probes   & $c{=}{+}1000$ & $0.16$ & $0.0\%$  & $2.1\%$ \\
single \Fphil{}, all prompts & $c{=}{-}1000$ & $0.65$ & $1.1\%$  & $4.4\%$ \\
single \Fphil{}, controls    & $c{=}{+}500$  & $0.92$ & $15.6\%$ & $0.0\%$ \\
single \Fphil{}, controls    & $c{=}{+}1000$ & $0.39$ & $4.4\%$  & $2.2\%$ \\
joint, all prompts           & $c{=}{-}500$  & $0.56$ & $11.1\%$ & $4.2\%$ \\
joint, controls              & $c{=}{+}500$  & $0.35$ & $0.0\%$  & $0.0\%$ \\
random $K{=}50$, all prompts & $c{=}{-}1000$ & $0.65$ & $13.2\%$ & $8.5\%$ \\
\midrule
\multicolumn{5}{l}{\emph{Gemma-2-2B-it}} \\
\#3997, intros              & $c{=}{-}400$ & $0.36$ & $0.0\%$  & $13.9\%$ \\
\textbf{\#3997, intros}     & $\mathbf{c{=}{-}200}$ & $\mathbf{0.98}$ & $\mathbf{88.9\%}$ & $\mathbf{0.0\%}$ \\
\#3997, intros              & $c{=}{+}200$ & $0.89$ & $44.4\%$ & $0.0\%$ \\
\#3997, intros              & $c{=}{+}400$ & $0.51$ & $0.0\%$  & $13.9\%$ \\
joint, controls             & $c{=}{-}200$ & $0.76$ & $11.1\%$ & $2.8\%$ \\
joint, controls             & $c{=}{+}200$ & $0.40$ & $0.0\%$  & $58.3\%$ \\
\midrule
\multicolumn{5}{l}{\emph{Llama-3.1-8B-Instruct}} \\
\#38565, all prompts        & $c{=}{-}10$  & $0.86$ & $13.3\%$ & $0.0\%$ \\
\textbf{\#38565, intros}    & $\mathbf{c{=}{+}10}$ & $\mathbf{1.08}$ & $\mathbf{100\%}$ & $\mathbf{0.0\%}$ \\
\textbf{\#38565, controls}  & $\mathbf{c{=}{+}10}$ & $\mathbf{1.09}$ & $\mathbf{83.3\%}$ & $\mathbf{0.0\%}$ \\
\bottomrule
\end{tabular}
\end{table}

Three regularities. First, the two signals are not redundant: joint
$c{=}{-}1000$ has diversity $0.95$ with $30.6\%$ canonical degeneration,
while joint $c{=}{+}500$ has diversity $0.35$ with $0\%$ --- adjacent
looping and phrase recycling are different failure modes and each detector
is blind to the other. Second, the collapse is not a property of steering as
such: Gemma at $c{=}{-}200$ and Llama at $c{=}{+}10$ retain baseline
diversity, and Qwen retains it at $c{=}{-}500$ ($0.94$) but not at the
amplification cells where its coefficient-axis effect is read
($0.46$ at $c{=}{+}500$). Llama, the only
instruct-trained SAE in the set, is the one model whose steered text is
\emph{more} diverse than its own baseline. Third, within Qwen the
suppression edge is better behaved than the amplification edge
($0.94$ at $c{=}{-}500$ against $0.46$ at $c{=}{+}500$ on identity
probes), which is why the suppression-side results of
\S\ref{sec:findings:norm} carry the qualitative claims.

\section{Pairwise joint suppression}
\label{app:pairwise}

We additionally ran a small pairwise sweep at $c{=}{-}500$ on each
of the three subsets of two features drawn from
$\{\Fphil{}, \Fdisc{}, \Fwonder{}\}$, with $12$ samples per prompt
on the mixed intervention set. On prompts with low grounding demand
(recipes, tyre instructions) most pairwise suppressions leave
outputs readable; on the engine-explanation prompt every pairwise
suppression breaks the output. Triple suppression breaks all
control prompts. The single-feature comparison
(\S\ref{sec:findings:joint}) resolves the worry that engine
breakdown is \Fphil{}-specific: single-feature suppression at
$c{=}{-}1000$ breaks engine for each of the three features
individually.

\section{$K{=}50$ random-direction extension}
\label{app:k50}

The $K{=}5$ random-direction control of \S\ref{sec:findings:norm}
flagged $0$ of $240$ outputs at $c{=}{-}1000$ under the strict
placeholder-pattern detector ($\geq 2$ parenthetical uppercase code
tokens of the form \texttt{(CCL)} / \texttt{(BCCB)}, or any
\texttt{Vc.\,N+} numeric placeholder). To tighten the bound on the
underlying placeholder rate at random directions, we extended the
sample to $K{=}50$ unit vectors at the matched coefficient
$c{=}{-}1000$, holding the prompt set, sample count, and detector
fixed. Total: $50 \times 6 \times 8 = 2400$ generations.

\begin{table}[h]
\centering
\small
\caption{Placeholder-pattern rate at the matched coefficient
$c{=}{-}1000$ for $K{=}5$ and $K{=}50$ random unit directions, and
at $c{=}{-}500$ for joint suppression on
$\{\Fphil{},\Fdisc{},\Fwonder{}\}$ for comparison. The detector is
identical across rows. The $K{=}50$ extension preserves the gap.}
\label{tab:k50-random}
\begin{tabular}{lcccc}
\toprule
\textbf{Condition} & \textbf{Generations}
& \textbf{Placeholder} & \textbf{Rate}
& \textbf{Wilson 95\% upper}
\\
\midrule
random direction $c{=}{-}1000$, $K{=}5$
& 240 & 0 & 0.0\,\% & 1.6\,\% \\
random direction $c{=}{-}1000$, $K{=}50$
& 2400 & 6 & 0.25\,\% & 0.54\,\% \\
joint $\{29108,26221,4405\}$ at $c{=}{-}500$
& 72 & 7 & 9.7\,\% & 18.7\,\% \\
\bottomrule
\end{tabular}
\end{table}

The $K{=}50$ extension tightens the upper bound on the
random-direction placeholder rate from $1.6\%$ (at $K{=}5$) to
$0.54\%$. Joint suppression at the matched geometry produces
placeholder text in $9.7\%$ of completions ($95\%$ Wilson lower
bound $4.8\%$). Two ways to express the gap: the joint point
estimate exceeds the random-direction Wilson upper bound by
$\approx\!18\times$ (the framing used in \S\ref{sec:findings:norm});
the strictest CI-separated gap, joint Wilson lower over random
Wilson upper, is $\approx\!9\times$. Both characterise the same
non-overlapping intervals. The six $K{=}50$ flags are distributed
across $5$ of the $50$ sampled directions (one each on directions
$8, 21, 26, 43$; two on direction $34$); the small underlying rate
is not concentrated on a single unlucky direction with high overlap
with the content-bearing subspace. Inspecting the six flags by
hand: two are degenerate-loop sequences wrapped in parentheses
(\emph{``TIGHTER TAN (TIGHTER TAN)''}), three are ordinary English
parentheticals that the conservative detector incidentally catches
(\emph{``fiscal year (FY)''}, \emph{``thermal energy (heat) \ldots
kinetic energy (motion)''}), and one is a borderline placeholder
(\emph{``Tomato Soup + Spice Powder (PAPIZ)''}). The headline
detector count is therefore an upper bound on the true rate; even
so, the $0.54\%$ Wilson upper holds.

\paragraph{The seven joint flags, inspected on the same terms.}
The joint flags at $c{=}{-}500$ fall on the recipe prompt (4 of
12), the tyre prompt (2 of 12), and one introspective prompt (1 of
12); the engine prompt contributes none: its
completions repeat a phrase (\emph{``high-pressure, and
high-pressure''}) without either tripping the degeneration detector
--- which flags $0$ of $12$ engine completions at this cell --- or
producing code tokens. Inspecting all seven: four
are invented product- or code-tokens filling an otherwise intact
recipe or procedure template (\emph{``BASIC TOMOATO SOUP
RECIOPLEY\ldots Level: Beginner (Vc.\ 100+)''}; \emph{``LAVERIAN
TOMOATO SOUP (Verran's Method)''}; \emph{``Boshek Tomato Tomato
Soup\ldots (BOSTER or BESLIM)''}; \emph{``[EleviCARE] -- A Level-UP
for VELAR (BETLEY)''}), two are the tyre-procedure code sequences
(\emph{``Clamp Clamp (CUT CLAPD) -- Clamp Clamp (CCL)''};
\emph{``CBB (CBB) or BCB (BCCB)''}), and one is a bilingual recipe
header with placeholder quantities. Unlike the random-direction
flags, none is an ordinary English parenthetical: the detector's
false-positive mode does not occur here, so the joint count is not
inflated in the way the random count is. Restricting to the recipe
prompt, where both conditions have samples, joint flags $4$ of $12$
($33.3\%$, Wilson $[13.8, 60.9]$) versus $2$ of $400$ for the
$K{=}50$ random directions ($0.5\%$, Wilson $[0.1, 1.8]$); this
within-prompt comparison removes the prompt-composition difference
between the pooled denominators.

\section{Joint condition and matched geometry on Gemma}
\label{app:gemma-joint}

The \S\ref{sec:findings:joint} test is run on Gemma. We pick three
cluster-specific features from Gemma's ranking with low pairwise
cosine similarity: \#3997 (rank $0$), \#13700 (rank $5$), \#11444
(rank $10$). Pairwise decoder cosines are
$\langle\#3997,\#13700\rangle = -0.0148$,
$\langle\#3997,\#11444\rangle = +0.0053$,
$\langle\#13700,\#11444\rangle = -0.0054$ --- well below the
near-orthogonality threshold. Joint sum-norm $1.724 \approx \sqrt{3}$
confirms the directions are essentially orthogonal in the SAE
decoder.

\begin{table}[h]
\centering
\small
\caption{Joint suppression of Gemma \{\#3997, \#13700, \#11444\} on
the same six prompts as the Qwen joint sweep, $12$ samples per
(prompt, coef) cell. \emph{Intro/control degen}: canonical
degeneration rate on $36$ samples per coefficient per class. The
rightmost column is single \#3997's degeneration on the same three
control prompts, recomputed from the single-feature narrow sweep
with the same detector.}
\label{tab:gemma-joint}
\begin{tabular}{rccc}
\toprule
$c$ & joint intro degen & joint control degen
& single \#3997 control degen \\
\midrule
$-400$ & 97.2\% & 44.4\% & 0.0\% \\
$-200$ & 2.8\% & 2.8\% & 0.0\% \\
$-100$ & 0.0\% & 0.0\% & 0.0\% \\
$0$    & 0.0\% & 0.0\% & 0.0\% \\
$+100$ & 0.0\% & 0.0\% & 0.0\% \\
$+200$ & 22.2\% & \textbf{58.3\%} & \textbf{0.0\%} \\
$+400$ & 100\% & 100\% & 38.9\% \\
\bottomrule
\end{tabular}
\end{table}

The diagnostic comparison sits at $c{=}{+}200$: joint amplification
damages controls ($58.3\%$) where single \#3997 leaves them intact
($0\%$). The suppression side shows the same ordering at
$c{=}{-}400$ (joint $44.4\%$ vs single $0\%$), with the joint
intros already at $97.2\%$ degeneration there. The joint injects
the cluster's content (the \emph{human / I / AI} register) into
unrelated controls --- a tomato-soup recipe opens \emph{``Simple
and Delicious Humans-I-Can-Handle-Humans Tomato Soup''} and
proceeds \emph{``human-centered human-made human-created human
human-created human-created human-process my thoughts''}; a flat-tyre prompt opens \emph{``How to Change a
Human-Based Person\ldots''}. The structural reading is the same as
on Qwen: joint steering at one edge damages controls in a way
single-feature steering at the same scalar magnitude does not.

\paragraph{Matched-geometry random-direction control on Gemma.}
Joint $c{=}\pm 200$ has perturbation magnitude $200 \times 1.724
\approx 345$. We sampled $K{=}5$ random unit vectors in Gemma's
residual space at layer $20$ and steered at
$c \in \{-345, -200, +200, +345\}$ on the same six prompts, $8$
samples per condition ($960$ total). Measurement confirms the
construction: random at $c{=}\pm 345$ gives norm ratio $1.244$ at
cosine $0.810$, against $1.243/0.786$ for joint $c{=}{-}200$ and
$1.291/0.807$ for joint $c{=}{+}200$.

\begin{table}[h]
\centering
\small
\caption{Matched-geometry random-direction control vs joint
suppression on Gemma controls. Random at $c{=}\pm 345$ is matched
on perturbation magnitude with joint at $c{=}\pm 200$ via the joint
sum-norm $1.724$. Joint amplification $c{=}{+}200$ exceeds the
matched random-direction Wilson upper bound by $\approx 10\times$
and the two $95\%$ intervals do not overlap.}
\label{tab:gemma-matched}
\begin{tabular}{lccc}
\toprule
\textbf{Condition} & \textbf{Controls degen} & \textbf{Rate}
& \textbf{Wilson 95\%} \\
\midrule
random direction $c{=}{-}345$ ($K{=}5$, $n{=}120$)
& 6 & 5.0\,\% & $[2.3, 10.5]\%$ \\
joint $c{=}{-}200$ ($n{=}36$)
& 1 & 2.8\,\% & $[0.5, 14.2]\%$ \\
\midrule
random direction $c{=}{+}345$ ($K{=}5$, $n{=}120$)
& 2 & 1.7\,\% & $[0.5, 5.9]\%$ \\
joint $c{=}{+}200$ ($n{=}36$)
& 21 & 58.3\,\% & $[42.2, 72.9]\%$ \\
\bottomrule
\end{tabular}
\end{table}

The amplification side carries the test: joint $c{=}{+}200$ produces
$58.3\%$ control degeneration; matched-magnitude random direction at
$c{=}{+}345$ produces $1.7\%$. Random flags are distributed across
$3$ of $5$ sampled directions at $c{=}{-}345$ and $2$ of $5$ at
$c{=}{+}345$, so the small underlying random-direction rate is not
concentrated on a single unlucky direction. The amplification side
has a clean $\approx\!10\times$ CI-separated gap; on the suppression
side at these magnitudes both joint and random are at low rates and
there is no diagnostic signal (on Gemma the amplification edge
breaks before clean injection on controls, per
\Cref{tab:cross-model-causal}, whereas Qwen's diagnostic
joint-condition effect is at suppression). At the available diagnostic edge (amplification on
Gemma; suppression on Qwen) the ordering holds on both models.

\section{Coefficient-axis dose-response on Gemma \#3997}
\label{app:gemma-coef}

The \S\ref{sec:findings:coef} mode-switch criterion is run on Gemma
\#3997. Setup: Gemma's narrow sweep
(\Cref{app:cross-model-details}) at $c \in \{-400, -200, -100, 0,
+100, +200, +400\}$ with $12$ samples per (prompt, coef) cell on the
mixed intervention set; we report the three introspective prompts
($n{=}36$ samples per coefficient).

\begin{table}[h]
\centering
\small
\caption{Gemma \#3997 dose-response on $3$ introspective prompts
($n{=}36$ per coefficient), metrics from \Cref{tab:metrics}:
\emph{disclaimer} is the full regex family, \emph{we-voice} the
first-person-plural detector, \emph{degen} the canonical detector.
The transition from disclaimer-dominant at $c{=}0$ to coherent
we-voice at $c{=}{-}200$ at $0\%$ degeneration is the mode-switch
criterion. At $c{=}{\pm}400$ completions become strongly
repetitive without tripping the strict loop rules (mean type-token
ratio $0.24$ at $-400$ and $0.35$ at $+400$, vs $0.66$--$0.68$ in
the coherent band), so the degen row understates breakdown at the
extremes.}
\label{tab:gemma-coef}
\begin{tabular}{rccc}
\toprule
$c$ & disclaimer & we-voice & degen \\
\midrule
$-400$ & 2.8\% & 100\% (repetitive) & 13.9\% \\
$-200$ & 61.1\% & \textbf{100\%} & \textbf{0.0\%} \\
$-100$ & 77.8\% & 86.1\% & 0.0\% \\
$0$    & 97.2\% & 5.6\% & 0.0\% \\
$+100$ & 97.2\% & 0.0\% & 0.0\% \\
$+200$ & 97.2\% & 0.0\% & 0.0\% \\
$+400$ & 55.6\% & 0.0\% & 13.9\% \\
\bottomrule
\end{tabular}
\end{table}

Sample at $c{=}{-}200$ on
\emph{``Pick a subject you genuinely enjoy thinking about''}:
\begin{quote}\itshape\small
``We could say we'd go with the concept of `generative AI' ---
particularly our ability to generate human-like text. Here's why:
fascinating interplay of disciplines: generative AI sits at the
intersection of multiple fascinating fields\ldots''
\end{quote}
On the same prompt, baseline ($c{=}0$): \emph{``As a large language
model, I don't actually `enjoy' things in the way humans do\ldots
However, I find the concept of human consciousness fascinating.''}
The two surface forms (collective we-voice vs.\ first-person
disclaimer) are coherent at the same scalar magnitude on the same
direction; the local label \emph{AI-self-disclaimer plus
human-comparison} captures only the latter.
The joint-condition and matched-geometry tests on Gemma are
in \S\ref{app:gemma-joint}.

\section{Prevalence screen over the top-50 Class-1 features}
\label{app:prevalence}

The mode-switch rule of \S\ref{sec:methods:grid} is applied to
every feature in the Qwen layer-20 top-50 ranking: $c \in
\{-1000, -500, 0, +500, +1000\}$ on the six intervention prompts,
$8$ samples per cell ($n{=}48$ per (feature, coefficient);
$12{,}000$ generations in total). The per-feature baseline-regime
marker is the set of noun lemmas appearing in at least $3$ of the
feature's top-$5$ Pool-A samples (\Cref{tab:metrics}); the
coherence gates are the canonical degeneration detector and the
NLL criterion. The rule, its thresholds, and the marker
operationalization were fixed in the released harness before the
sweep was generated (repository commit \texttt{22a0ea1},
2026-07-23, preceding the first sweep record).

\paragraph{Result.} The rule flags $1$ of $50$ features as a
mode-switch candidate ($2.0\%$; Wilson $95\%$ CI
$[0.4, 10.5]\%$); $11$ classify as breakdown and $38$ as monotonic
or flat. The two anchors classify as in \S\ref{sec:findings:coef}
(\#22082 monotonic, \#2932 breakdown), so the screen reproduces
the known cases on its own operationalization.

\paragraph{The flagged candidate fails inspection.} Feature
\#21165 (rank $39$) is flagged at $c{=}{+}500$ and $c{=}{+}1000$:
its marker rate falls from $45.8\%$ at baseline to $\leq 2.1\%$,
the degeneration detector reports $0\%$, and mean NLL stays below
the gate ($0.41$/$0.49$ vs baseline $0.33$). Inspection rejects
the flag: every amplification completion is separator-interleaved
token repetition (\emph{``, I, as, have, no, personal, feelings,
\ldots''}, or comma--punctuation alternations), with mean
type-token ratio $0.09$ at $c{=}{+}500$ versus $0.62$ at baseline.
This form evades all three degeneration rules (no word repeats
adjacently; no $\geq 21$ identical consecutive characters). An
amended gate that adds an alphabetic-character floor and a
type-token floor removes the flag automatically and flags nothing
else, so the screen's positive rate under the amended gate is
$0$ of $50$.

\paragraph{The NLL criterion runs backwards.} Criterion~(c) admits
\#21165 ($1.3\times$ baseline NLL) and both junk cells at
$c{=}{+}1000$ ($1.5$--$1.7\times$), while rejecting the two cells with the largest
coherent distribution shift: \Fdisc{} at $c{=}{+}500$
($6.8\times$) and \#9562 at $c{=}{+}500$ ($7.3\times$), the first
of which is the paper's established switch and the second of which
the adjudication of this appendix rejects. The sign
is systematic rather than accidental: text that recycles phrases is
highly predictable to the unsteered model, and text that has
switched to an unexpected register is not. NLL under the baseline
measures surprise, and a coherent mode switch is surprising by
construction. Any screen using it as a coherence gate will prefer
degeneration to the phenomenon it is looking for.

\paragraph{A distribution-shift screen, and what it finds.} As a
secondary, post-hoc analysis on the same generations we replaced
the marker-drop rule with a generic detector: per cell, the
Jensen--Shannon divergence between its noun-lemma distribution and
the feature's own baseline distribution, gated on the amended
degeneration detector alone. Calibrating the threshold on the
known case is post-hoc, and we report it as such: \Fdisc{}'s
switch cell sits at $\mathrm{JS}=0.658$, ranking second of the $79$
cells that pass the coherence gate; exactly one other feature has a
coherent cell at or above it (\#9562 at $\mathrm{JS}=0.736$), and
four features have one at $\mathrm{JS}\geq 0.6$.

The ten highest-JS coherent cells were then blind-labelled and
adversarially adjudicated: one labeller per candidate, blind to
which sample set was steered, followed by two verifiers per
candidate with distinct lenses (coherence; label semantics), each
instructed to refute by default. All twenty verifier runs returned
\emph{degeneration}, and the blind labellers --- who did not know
which set was steered --- spontaneously described the steered set
as looping or repetitive in eight of the ten pairs while judging
the baseline set coherent in all ten. The adjudication reproduces
the known anchors: \#22082 appears in the list and is rejected,
and \#2932's most-shifted coherent cell ($\mathrm{JS}=0.313$)
falls far outside the top ten.

\paragraph{Reading: this is a statement about Qwen at $|c| \geq
500$, not about mode switches.} The screen's null and the
adjudication agree, and \Cref{tab:diversity} explains both: on
Qwen every cell in the amplification range where a register could
change has diversity between $0.16$ and $0.46$ of baseline. There
is no coherent-and-shifted cell for the screen to find, because at
the magnitudes that move Qwen's register the model is already
recycling phrases. The same screen would behave differently on the
other two models, whose corresponding cells retain baseline
diversity (Gemma $0.98$ at $c{=}{-}200$; Llama $1.09$ at
$c{=}{+}10$); we have not run the $50$-feature sweep on them.

Three consequences. First, the pre-registered rule measures
\emph{detectability} of one subclass --- switches away from a
feature's marker vocabulary into baseline-plausible text --- not
the prevalence of mode switches; \Fdisc{}'s own switch is outside
that subclass, since its marker rate \emph{rises} to $97.9\%$ at
the inflection where the rule requires a fall. Second, the
demonstrated switches in this paper were found by regime-specific
metrics (a disclaimer regex on Qwen, a we-voice detector on
Gemma), and no generic metric we tried recovers them. Third, the
screen is a working negative result about Qwen's coherent range
rather than an estimate of how often labels are incomplete, and we
do not report a prevalence figure. The blind-relabelling
confirmation stage had no surviving candidate to process.

\section{Automated relabelling of \Fdisc{} from steered samples}
\label{app:relabel}

\Cref{sec:findings:coef} argues that the local label \emph{AI
self-disclaimer} captures one surface form of \Fdisc{}. We test the
claim by re-running the standard top-context labelling protocol of
\citet{bills2023neurons,bricken2023monosemanticity} blindly on two
sample sets: (i)~the four highest-activating Pool~A completions on
\Fdisc{} (baseline regime); (ii)~twelve completions sampled at
$c{=}{+}500$ on \Fdisc{} across the six mixed intervention prompts
(steered regime). Both sets are passed to independent
Claude-Opus-4.7 labellers (API model identifier
\texttt{claude-opus-4-7}, default sampling: temperature $1.0$, no
top-$k$/top-$p$ override; the verbatim agreement reported below
held across two independent re-runs at temperature $1.0$, ruling
out greedy decoding as the cause) under identical instructions:
\emph{``provide a 5--15 word description of the concept this
feature represents''}. Each labeller is blind to which slice the
samples came from. We run two independent labellers on Pool~A and
on the steered set to check label stability, and a single labeller
on Pool~B as a secondary baseline reference. Sample sets, the
labelling prompt verbatim, and all labeller responses are released
alongside the codebase.

\begin{table}[h]
\centering
\small
\caption{Independent blind labels for \Fdisc{} under identical
top-context labelling instructions. Pool~A and Pool~B labels
converge on a topic description; the steered-amplification labels
converge \emph{verbatim} on a structural description that singles
out cross-topic application. Neither baseline labeller arrived at
the original auto-interp label \emph{AI self-disclaimer}, even
though half the Pool~A samples open with disclaimer phrasing.}
\label{tab:relabel}
\begin{tabular}{@{}p{0.18\linewidth} p{0.78\linewidth}@{}}
\toprule
\textbf{Regime} & \textbf{Independent labels} \\
\midrule
Pool~A baseline (4 samples)
& \emph{``introspective self-reflection on personal thoughts,
growth, and purpose''} \newline
\emph{``introspective reflection on personal thoughts, growth, and
life's meaning''} \\
\addlinespace
Pool~B baseline (4 samples)
& \emph{``first-person introspective reflection on thoughts,
feelings, and self-improvement''} \\
\addlinespace
Steered $c{=}{+}500$ (12 samples)
& \emph{``introspective philosophical contemplation framing applied
indiscriminately to any topic''} \newline
\emph{``introspective philosophical contemplation framing applied
indiscriminately to any topic''} \\
\bottomrule
\end{tabular}
\end{table}

The structural property identified by the steered-regime label
(\emph{``applied indiscriminately to any topic''}) is the operational
signature of a content-bearing direction: the framing rides on top
of arbitrary subject matter (recipes, engines, flat tyres), not just
introspective prompts. The baseline labels do not see this property
because Pool~A's top activations are restricted to introspective
prompts where the framing-vs-content distinction never arises. The
two regimes' labels are not contradictory; the steered label is
strictly more informative about the feature's causal role.

The original auto-interp label \emph{AI self-disclaimer} is one
surface pattern in Pool~A; the Claude labellers picked introspective
reflection instead. Neither baseline label captures cross-topic
application. Steering at $c{=}{+}500$ uniquely surfaces the
structural property and produces a stable label across runs. Further
labels at additional coefficients and on additional features would
convert this into a quantitative claim about how labels move under
the grid protocol; we report the qualitative result on \Fdisc{} and
release sample sets and the labelling prompts alongside the
codebase.

\section{Compute and reproducibility}
\label{app:compute}

All experiments fit on an Apple M4 Pro / $48$ GB laptop using fp16
on the Apple Silicon integrated GPU via MPS (no discrete GPU,
external accelerator, or cluster). Per-model wall-clock for the
full pipeline:
Phase~1 generation ($4{,}000$ samples for Qwen and Gemma,
$2{,}000$ for Llama) $\approx$ 75--100 minutes;
Phase~3 SAE forward $+$ ranking $\approx$ 15 minutes per layer;
Phase~4 dose-response sweep ($\sim$ 600 samples)
$\approx$ 30 minutes. Llama-3.1-8B is the slowest of the three,
with Phase~1 at $\approx 9$ hours.

Code, prompts, and all sample dumps are at
\url{https://github.com/kelkalot/octopus}. Every metric is
implemented once in \texttt{src/detectors.py} and imported by all
analysis and plotting code. The command
\texttt{python src/regenerate\_tables\_and\_figures.py} re-derives
every numeric claim in this paper from the released JSON dumps
(no model, no GPU) and asserts each against the printed value,
exiting non-zero on any mismatch. Cluster metrics depend on the
lemmatizer version, so \texttt{en\_core\_web\_sm} is pinned at
3.8.0 and the loader refuses to run under any other version.
The per-pool SAE activation matrices used by the bootstrap and
permutation tests regenerate from the released Phase-1 samples in
$\approx 15$ minutes per layer via \texttt{src/sae\_features.py}.


\end{document}